\documentclass[preprint,12pt]{elsarticle}

\usepackage{amssymb}
\usepackage{amsmath}
\usepackage{url}
\usepackage[hidelinks]{hyperref}
\usepackage[utf8]{inputenc}
\usepackage{amsmath}
\usepackage{amsthm}
\usepackage{booktabs}
\usepackage{algorithm}
\usepackage[switch]{lineno}
\usepackage{wrapfig}
\usepackage{subcaption}
\usepackage{xcolor}
\usepackage{multirow}
\usepackage{algorithmic}
\newcommand{\mathbold}[1]{\ensuremath{\boldsymbol{\mathbf{#1}}}}

\newcommand{\g}{\,|\,}

\newcommand{\nestedmathbold}[1]{{\mathbold{#1}}}

\newcommand{\mbr}{\nestedmathbold{r}}

\newcommand{\mbt}{\nestedmathbold{t}}
\newcommand{\mbu}{\nestedmathbold{u}}

\newcommand{\mbw}{\nestedmathbold{w}}
\newcommand{\mbx}{\nestedmathbold{x}}

\newcommand{\mbz}{\nestedmathbold{z}}

\newcommand{\mbI}{\nestedmathbold{I}}

\newcommand{\mbdelta}{\nestedmathbold{\delta}}
\newcommand{\mbepsilon}{\nestedmathbold{\epsilon}}

\newcommand{\mbmu}{\nestedmathbold{\mu}}
\newcommand{\mbnu}{\nestedmathbold{\nu}}

\newcommand{\mbphi}{\nestedmathbold{\phi}}

\newcommand{\mbsigma}{\nestedmathbold{\sigma}}

\newcommand{\mbtheta}{\nestedmathbold{\theta}}

\newcommand{\mbxi}{\nestedmathbold{\xi}}

\newcommand{\mbPi}{\nestedmathbold{\Pi}}

\newcommand{\ELBO}{\textsc{elbo}}

\DeclareRobustCommand{\KL}[2]{\ensuremath{\textsc{kl}\left[#1\;\|\;#2\right]}}
\DeclareRobustCommand{\DV}[2]{\ensuremath{\textsc{dv}\left[#1\;\|\;#2\right]}}

\newcommand{\diag}{\textrm{diag}}

\newcommand{\cN}{\mathcal{N}}

\usepackage{amsmath,amsfonts,bm}

\def\1{\bm{1}}

\DeclareMathAlphabet{\mathsfit}{\encodingdefault}{\sfdefault}{m}{sl}
\SetMathAlphabet{\mathsfit}{bold}{\encodingdefault}{\sfdefault}{bx}{n}

\newcommand{\E}{\mathbb{E}}

\DeclareMathOperator*{\argmin}{arg\,min}

\journal{Artificial Intelligence}

\newtheorem{proposition}{Proposition}
\newtheorem{theorem}{Theorem}[section]
\newtheorem{property}[theorem]{Property}

\begin{document}

\begin{frontmatter}



\title{Federated Neural Nonparametric Point Processes}

\author[label1]{Hui Chen}
\author[label1]{Xuhui Fan}
\author[label2]{Hengyu Liu}
\author[label3]{Yaqiong Li}
\author[label1]{Zhilin Zhao}
\author[label4]{Feng Zhou}
\author[label5]{Christopher John Quinn}
\author[label1]{Longbing Cao}
\affiliation[label1]{organization={Macquarie University},
            city={Sydney},
            country={Australia}}

\affiliation[label2]{organization={Aalborg University},
            city={Aalborg},
            country={Denmark}}

\affiliation[label3]{organization={Australian Federal Government},
            city={Sydney},
            country={Australia}}

\affiliation[label4]{organization={Renmin University of China},
            city={Beijing},
            country={China}}

\affiliation[label5]{organization={Iowa State University},
            city={Ames},
            country={USA}}


\begin{abstract}
{Temporal point processes (TPPs) are effective for modeling event occurrences over time, but they struggle with sparse and uncertain events in federated systems, where privacy is a major concern. To address this, we propose \textit{FedPP}, a Federated neural nonparametric Point Process model. FedPP integrates neural embeddings into Sigmoidal Gaussian Cox Processes (SGCPs) on the client side, which is a flexible and expressive class of TPPs, allowing it to generate highly flexible intensity functions that capture client-specific event dynamics and uncertainties while efficiently summarizing historical records. For global aggregation, FedPP introduces a divergence-based mechanism that communicates the distributions of SGCPs' kernel hyperparameters between the server and clients, while keeping client-specific parameters local to ensure privacy and personalization. FedPP effectively captures event uncertainty and sparsity, and extensive experiments demonstrate its superior performance in federated settings, particularly with KL divergence and Wasserstein distance-based global aggregation.}
\end{abstract}


\begin{highlights}
\item We introduce FedPP, the first federated adaptation of Temporal Point Process (TPP) models, bridging the gap between TPPs and Federated Learning (FL) while addressing key challenges such as event sparsity, uncertainty, and privacy concerns.
\item We propose an innovative integration of neural embedding techniques within the kernels of Sigmoidal Gaussian Cox Processes (SGCPs), significantly enhancing their expressiveness and enabling effective utilization of historical data.
\item We develop a divergence-guided global aggregation mechanism, facilitating the secure sharing of neural embedding distributions between the server and clients, ensuring robust global modeling while preserving client-specific privacy.
\item Our method outperforms existing approaches on benchmark datasets, capturing event sparsity and uncertainty in federated environments without compromising privacy.
\end{highlights}


\end{frontmatter}


\section{Introduction}
{Temporal point processes (TPPs)~\cite{shchur2021neural,lin2021empirical,wang2024learning,li2024dual} have shown great promise in modeling the occurrences of events over a specific observation period. TPPs have been successfully applied to various domains, including neuroscience~\cite{linderman2016bayesian,apostolopoulou2019mutually}, where they are used to study neural firing patterns, healthcare~\cite{yan2019modeling,enguehard2020neural} for predicting medical incidents or patient admissions, finance~\cite{hawkes2018hawkes,fan2023forecasting} for market event analysis, and mobile app-based ride-hailing~\cite{du2016recurrent,okawa2019deep} for predicting transport demand. These applications highlight the flexibility and broad utility of TPPs in modeling event-driven data across diverse fields.}

Despite their utility, traditional TPPs methods face significant challenges when dealing with event data that is sparse, uncertain, and distributed in federated systems. For instance, in the context of sequential shopping behaviors, event sequences are often private, and aggregating all these sequences may lead to privacy risks. Moreover, the current TPPs models are usually centralized and not client-specific, making them unable to capture the sparsity and uncertainty of events in each client. In other words, privacy-preserving methods capable of modeling the inherent complexity of event data are needed in practical situations. 

{Federated learning (FL)~\cite{mcmahan2017communication,konevcny2016federated,hu2023privacy,du2024unified} offers a promising solution for addressing privacy concerns by decentralizing model training across clients. However, to the best of our knowledge, no existing work has explored federated learning in the context of TPPs, particularly with challenges related to event sparsity and uncertainty. This raises key research questions: How can we model client-specific event sequences while preserving privacy? How can we aggregate knowledge across clients without centralizing sensitive event data?}

In this paper, we focus on addressing these challenges by leveraging Sigmoidal Gaussian Cox Processes (SGCPs)~\cite{donner2018efficient,zhou2019scalable,zhou2020efficient}, which has been successfully used as a powerful TPPs models. Following typical TPPs models, SGCPs model an intensity function over the observation period. In particular, the events' intensities are learned by using a Gaussian process, which allows for capturing complex temporal patterns, while a sigmoidal transformation ensures valid non-negative intensity values. However, SGCPs are limited by the expressiveness of their kernel functions, particularly in capturing the nuances of event sparsity and uncertainty across clients and in effectively using historical information. 

To overcome these limitations, we propose FedPP, a Federated neural nonparametric Point Process model, which integrates neural embedding methods~\cite{du2016recurrent,zuo2020transformer,zhang2020self} into SGCP kernels to improve their flexibility and expressiveness, and operates by decentralizing the training process across clients to preserve privacy. On the \emph{client side}, the integrated neural embeddings may effectively summarize historical event information and generate flexible client-specific intensity functions. On the \emph{server side}, FedPP develops a novel divergence-guided global aggregation mechanism that communicates distributions of neural embeddings rather than individual parameters. This ensures common event information and uncertainties are shared in a privacy-preserving manner. By aligning the global distribution with client-specific distributions, FedPP personalizes the model for each client while maintaining a strong global performance. This structure effectively addresses both the privacy and modeling challenges associated with federated TPPs.

In summary, we make the following contributions:
\begin{itemize}
    \item We propose FedPP, the first federated version of Temporal Point Process models, which bridges the gap between temporal point processes (TPPs) and federated learning (FL), addressing the challenges of event sparsity, uncertainty, and privacy concerns.
    \item We introduce a novel integration of neural embedding methods into the kernels of Sigmoidal Gaussian Cox Processes (SGCPs), significantly improving their expressiveness and enabling effective utilization of historical information.
    \item We develop a divergence-guided global aggregation mechanism that shares distributions of neural embeddings between the server and clients, allowing for robust global modeling while preserving client-specific privacy.
    \item Our method demonstrates superior performance on benchmark datasets, effectively capturing event sparsity and uncertainty in federated settings without compromising privacy.
\end{itemize}

\section{Background}
Temporal point processes (TPPs) provide a fundamental framework for modeling event occurrences over time, with a broad range of applications in fields such as healthcare, finance, and social networks. Several advanced methods have been developed to improve the expressiveness of TPPs, including Sigmoidal Gaussian Cox Processes (SGCPs) for nonparametric modeling and neural embedding methods for capturing historical event dependencies. In this section, we provide an overview of TPPs, introduce SGCPs as a key nonparametric approach, and discuss neural embedding techniques that leverage historical event data to improve event prediction.

\subsection{Temporal Point Processes~(TPPs)} \cite{daley2007introduction} provided a probabilistic framework for modeling a set of random events over a time period. In TPPs, the rate of event occurrences is characterized by an intensity function $\lambda(t)$:
\begin{equation}
\lambda(t)=\lim _{\Delta t \rightarrow 0^{+}} \frac{\mathbb{E}[\mathcal{N}(t+\Delta t)-\mathcal{N}(t)]}{\Delta t},
\end{equation}
where $\mathcal{N}(t)$ denotes the number of events occurring up to time $t$. The intensity function $\lambda(t)$ is a non-negative function of time, indicating the likelihood of event occurrences at time $t$. Larger values of $\lambda(t)$ imply a higher chance of event occurrence. Given a set of $N$ observed event times $\mbt = \{t_1, \ldots, t_N\}$ with $t_i \in [0, T]$, the likelihood of observing this sequence in a point process is:
\begin{equation} \label{eq:intensity_function}
    \mathcal{L}\left(\lambda(t)|\{t_1, \ldots, t_N\}\right)=e^{-\int_{0}^T\lambda(t)dt}\prod_{i=1}^N\lambda(t_i).
\end{equation}

\subsection{Sigmoidal Gaussian Cox Processes (SGCPs)} SGCPs~\cite{donner2018efficient,zhou2019scalable,zhou2020efficient} is a powerful nonparametric framework for learning event data. In SGCPs, the intensity function $\lambda(t)$ is modeled as a sigmoidal transformation of a random function drawn from a Gaussian process, expressed as $\lambda(t)=m\cdot\sigma(f(t))$, where $f(\cdot)\sim\mathcal{GP}(\nu(\cdot), \kappa_{\cdot, \cdot;\mbw})$, and $\sigma(\cdot)=1/(1+\exp(-\cdot))$ ensures the intensity is within $(0,1)$. Here, $m$ is a scaling parameter, $f(t)$ is the random function, $\nu(t)$ is the mean function, and $\kappa_{t, t';\mbw}$ is the kernel function with hyperparameters $\mbw$. SGCPs are Bayesian nonparametric models capable of representing highly flexible intensity functions. However, traditional SGCPs rely on standard kernels like the RBF kernel, which may be inefficient to generate highly flexible intensity functions. Further, it might be difficult to directly use the rich historical information to predict future events for SGCPs.

\subsection{Neural Embedding Methods for Event Data}
For a given event $t_i$, its history, denoted as $\mathcal{H}_{t_i}=\{t_{i'}: t_{i'}<t_i\}$, can significantly influence the occurrence of the next event. Neural embedding methods~\cite{du2016recurrent,zuo2020transformer,zhang2020self} encode this history into a vector $\mathbf{h}_{i-1}$, referred to as the historical embedding. This embedding can be used to parameterize the conditional distribution of the next event time $t_{i+1}$:
\begin{equation} \label{eq:neural-embedding}
t_{i+1} \sim P_{\theta}(t_{i+1}|\mathbf{h}_{i}), \ \ \ \text{where}\ \ \ \mathbf{h}_{i}=f_{\mathrm{update}}(\mathbf{h}_{i-1},\mathbf{e}_i).
\end{equation}
Here, $f_{\mathrm{update}}$ represents a recurrent update function such as RNN, LSTM, or an attention-based layer, and $\mathbf{e}_i$ is the embedding of event $t_i$~\cite{xue2023easytpp}. Neural embedding methods are effective in summarizing historical event information. However, many of these methods adopt an autoregressive strategy, predicting event times based on previous events, rather than learning the intensity function directly from the TPPs perspective, which may lead to a lack of clear statistical understanding of event dynamics over time.

\section{The FedPP Methodology}
In this section, we introduce the methodology behind our Federated neural nonparametric Point Process model (FedPP). FedPP is designed to address the challenges of modeling event sparsity and uncertainty in federated systems while preserving data privacy. The main idea of FedPP is to integrate neural embeddings with Sigmoidal Gaussian Cox Processes (SGCPs) on the client side to model client-specific event dynamics and uncertainties while effectively summarizing historical information. It also introduces a divergence-based mechanism for sharing kernel hyperparameters between the server and clients, while keeping client-specific parameters local to preserve privacy and support personalization. This section is structured as follows:
\begin{itemize}
    \item \textbf{Client-side modeling} (Section~\ref{sec:alg1}): We describe how each client uses neural SGCPs to model its event sequence, including the use of neural embeddings to summarize historical information, and sparse Gaussian processes to reduce computational complexity.
    \item \textbf{Server-side aggregation} (Section~\ref{sec:alg2}): When targeting at aggregating parameters' distributions, we further introduce a divergence-guided global aggregation mechanism that updates global distribution based on the clients' variational distributions.
    \item \textbf{Bi-level optimization} (Section~\ref{sec:alg3}): We formalize the overall optimization framework for FedPP as a bi-level problem, combining local and global updates.
    \item \textbf{Detailed Algorithm Steps} (Section~\ref{sec:alg4}): We present the FedPP algorithm in a practical federated learning setting, outlining the three key steps—computing expected log-likelihood, applying mean-field variational inference, and optimizing the local objective function using first-order methods. 
\end{itemize}

\subsection{Client-side modeling: Neural SGCPs}\label{sec:alg1}
{Each client in the federated learning (FL) framework models its observed event sequence $\mbt_c = \{t_{c,i}\}_{i=1}^{n_c}$, where $n_c$ is the number of events in client $c$, using a client-specific SGCP. The intensity function $\lambda_c(t)$ is expressed as:
\begin{equation}
\lambda_c(t) = m_c \cdot \sigma(f_c(t)), \quad f_c(t) \sim \mathcal{GP}(\nu_c, \kappa_{\cdot, \cdot; \mbw_c}),
\end{equation}
where $\mathcal{GP}(\nu_c, \kappa_{\cdot, \cdot; \mbw_c})$ is a Gaussian process with client-specific mean $\nu_c$ and kernel $\kappa_{\cdot, \cdot; \mbw_c}$. In our neural SGCPs for each client $c$, neural embedding methods are used to encode the historical information of events and improve the expressiveness of the intensity function.}

{\noindent\textbf{Neural Embedding.} For each event $t_i$, we encode its history $\mathcal{H}_{t_i} = \{t_{i'}: t_{i'} < t_i\}$ using a neural embedding method, by following Eq.~(\ref{eq:neural-embedding}). We propose that the resulting embedding $\mathbf{h}_{i}$ can be used to compute the kernel values between two time points $t_i$ and $t_j$ as:
\begin{equation}
\kappa_{\mathbf{t}_i, \mathbf{t}_j; [\tilde{\mbw}; r; l]} = r \cdot \exp\left(-\frac{\|\mathbf{h}_i - \mathbf{h}_j\|^2}{2l^2}\right),
\end{equation}
where $\tilde{\mbw}$ are the parameters of the neural network forming the historical embedding, $r$ is the scaling parameter and $l$ is the length-scale parameter for the RBF kernel. By incorporating neural embeddings into the kernel of the SGCP, we efficiently summarize historical event information and generate a more expressive intensity function. This addresses the traditional kernel limitations in SGCPs.}

{\noindent\textbf{Prior and Variational Distributions.} FedPP sets the prior distribution for the parameter vector $\mbw = [\tilde{\mbw}; \log r; \log l]$, which is a combination of the neural network parameters $\tilde{\mbw}$ and the RBF kernel parameters $\log r$ and $\log l$, as an isotropic multivariate Gaussian distribution: 
\begin{equation}
p_{\mbtheta}(\mbw_c) = \mathcal{N}(\mbw_c; \mbmu, \operatorname{diag}(\mbsigma^2)),
\end{equation}
where $\mbtheta = [\mbmu; \mbsigma^2]$, with $\mbmu \in \mathbb{R}^{(M_{\mbw} + 2) \times 1}$ and $\mbsigma^2 \in [\mathbb{R}^+]^{(M_{\mbw} + 2) \times 1}$. This prior distribution is shared across all clients for their respective parameter weights ${\mbw_c}$. Each client $c$ has its own variational distribution, denoted as $q_{\mbphi_c}(\mbw_c) = \mathcal{N}(\mbw_c; \mbr_c, \operatorname{diag}(\mbdelta_c^2))$, where the variational parameters $\mbphi_c = [\mbr_c; \mbdelta_c^2]$ are client-specific, with $\mbr_c \in \mathbb{R}^{(M_{\mbw} + 2) \times 1}$ and $\mbdelta_c^2 \in [\mathbb{R}^+]^{(M_{\mbw} + 2) \times 1}$.}

{\noindent\textbf{Sparse Gaussian Processes.} The sparse Gaussian process method is employed for each client to reduce the computational cost of neural SGCPs from $\mathcal{O}(n_c^3)$ to $\mathcal{O}(n_cM^2)$, where $M$ is the number of inducing points ${\mbu}_c \in \mathbb{R}^{M \times 1}$. By fixing the inducing locations ${\mbz} \in \mathbb{R}^{M \times 1}$ as constants that are equally distributed across the time period for all clients, the generative process of the neural SGCP for client $c$ can be approximated as follows:
\begin{align}
{\mbw}_c &\sim p_{\mbtheta}({\mbw}_c), \quad {\mbu}_c \sim \mathcal{N}(\mbu_c; \nu_c, \kappa_{\mbz,\mbz; \mbw_c}), \label{eq:prior-of-u}\\
\tilde{f}_c(\cdot) &\sim \mathcal{N}(\tilde{f}_c(\cdot); \nu_c + \kappa_{\cdot, \mbz; \mbw_c}\kappa_{\mbz, \mbz; \mbw_c}^{-1}({\mbu}_c - \nu_c), \kappa_{\cdot,\cdot; \mbw_c} - \kappa_{\cdot, \mbz; \mbw_c}\kappa_{\mbz, \mbz; \mbw_c}^{-1}\kappa_{\mbz, \cdot; \mbw_c}), \label{eq:prior-of-f}\\
\mbt_c &\sim \text{PoissonProcess}(m_c \cdot \sigma(\tilde{f}_c(\cdot))). \label{eq:prior-of-t}
\end{align}
Eq.~\eqref{eq:prior-of-u} describes the prior distribution of the inducing points ${\mbu}_c$, while Eq.~\eqref{eq:prior-of-f} represents the prior distribution of the approximated function $\tilde{f}_c(\cdot)$. Finally, Eq.~\eqref{eq:prior-of-t} describes the Poisson process with the intensity function $m_c \cdot \sigma(\tilde{f}_c(\cdot))$. These approximations $\{\tilde{f}_c(\cdot)\}_{c=1}^C$ are direct results of the sparse Gaussian process method.}

\subsection{Server-side aggregation: Divergence-guided global aggregation}\label{sec:alg2}
In FedPP, the prior distribution $p_{\mbtheta}(\cdot)$, specifically its parameter $\mbtheta$ which is shared among all the clients, is maintained on the server. During each communication round, the server broadcasts the prior distribution $p_{\mbtheta}(\cdot)$ to all participating clients. After each client updates its variational distribution $q_{\mbphi_c}(\mbw_c)$ independently, the variational parameters $\mbphi_c$ are uploaded to the server. The global aggregation is performed by minimizing the sum of divergences between $q_{\mbphi_c}(\mbw_c)$ and $p_{\mbtheta}({\mbw}_c)$ with respect to $\mbtheta$, expressed as:
\begin{equation}
\label{eq:target_global}
\min_{\mbtheta} \sum_{c=1}^C \DV{q_{\mbphi_c}({\mbw}_c)}{p_{\mbtheta}({\mbw}_c)},
\end{equation}
where $\DV{\cdot}{\cdot}$ represents a divergence between two probability distributions. This process finds the optimal prior distribution $p_{\mbtheta^*}(\cdot)$ that is closest to all the variational distributions $\{q_{\mbphi_c}(\mbw_c)\}_{c=1}^C$ of clients.

\begin{figure}[t]
\centering
\includegraphics[width=1\linewidth]{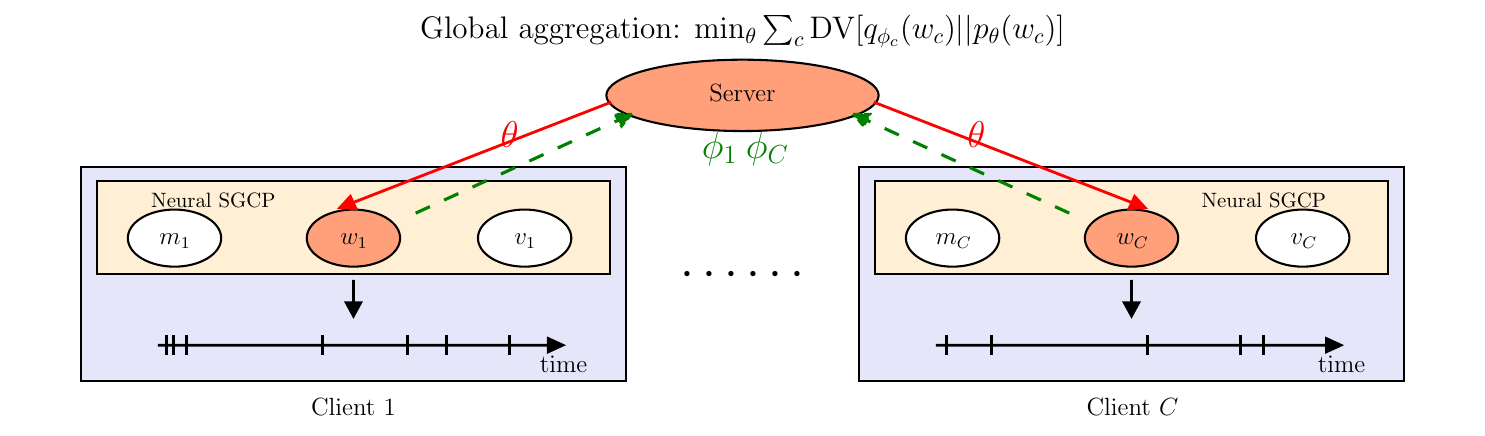}
\caption{Federated point process communication between server and clients. The variational distribution $q_{\mbphi_c}(\mbw_c)$ of clients are updated to the server, while $m_c$ and $\nu_c$ are kept at the clients. The global aggregation optimizes $\mbtheta$ to minimize the sum of divergences between the variational distribution $q_{\mbphi_c}(\mbw_c)$ of clients and its prior $p_{\mbtheta}(\mbw_c)$. The server then broadcasts the optimized $\mbtheta$ value to all the clients.}
\label{fig:rough-visualisation}
\end{figure}

Our proposed divergence-guided global aggregation method extends beyond temporal point processes (TPPs). This approach can be applied in any federated learning scenario where probability distributions are communicated between the server and clients. Fig.~\ref{fig:rough-visualisation} illustrates how $\mbtheta$ is updated for the prior distribution $p_{\mbtheta}(\mbw_c)$.


To solve Eq.~\eqref{eq:target_global}, we consider several choices for $\DV{\cdot}{\cdot}$, primarily KL Divergence and Wasserstein Distance. There are several ways to define $\DV{\cdot}{\cdot}$, each leading to different global aggregation rules:

{\noindent\textbf{(1) KL Divergence.} One of the most widely used divergences is the KL divergence. The sum of KL divergences:
\begin{equation}
\sum_{c=1}^C \KL{q_{\mbphi_c}({\mbw}_c)}{p_{\mbtheta}({\mbw}_c)}
\end{equation}
reaches its minimum for $\mbtheta = \{\mbmu, \mbsigma^2\}$ when:
\begin{equation}
\mbmu = \frac{1}{S} \sum_{c \in \mathbb{S}} \mbr_c, \quad \mbsigma^2 = \frac{1}{S} \sum_{c \in \mathbb{S}} \left[\mbdelta_c^2 + \mbr_c^2 - \mbmu^2\right],
\end{equation}
where $\mbphi_c = \{\mbr_c, \mbdelta_c^2\}$ and $\mathbb{S}$ is a random set of clients of size $S$. Unlike the FedAvg mechanism \cite{mcmahan2017communication,chen2024fedsi}, which averages $\{\mbr_c\}_{c=1}^C$ and $\{\mbdelta_c^2\}_{c=1}^C$ for $\mbmu$ and $\mbsigma^2$, our FedPP method incorporates the sample variance of $\{\mbmu_c\}_{c=1}^C$, which is $\sum_{c \in \mathbb{S}} [\mbmu_c^2 - \mbmu^2]/{S}$. This adjustment is essential because the spatial distribution of $\{\mbmu_c\}_{c=1}^C$ can have non-negligible variance.}

{\noindent\textbf{(2) Wasserstein Distance.} Compared to KL divergence, Wasserstein distance is more robust, especially when the supports of two distributions differ. The sum of Wasserstein distances:
\begin{equation}
\sum_{c=1}^C W_2[q_{\mbphi_c}({\mbw}_c) \| p_{\mbtheta}({\mbw}_c)]
\end{equation}
reaches its minimum for $\mbtheta = \{\mbmu, \mbsigma^2\}$ when:
\begin{equation}
\label{eq:wasserstein-distance}
\mbmu = \frac{1}{S} \sum_{c \in \mathbb{S}} \mbr_c, \quad \mbsigma = \frac{1}{S} \sum_{c \in \mathbb{S}} \mbdelta_c.
\end{equation}
In this case, the Wasserstein distance computes the average of the standard deviations of clients, yielding a more conservative estimate compared to FedAvg and KL divergence.}

\noindent\textbf{(3) Other Divergence Measures.} When closed-form solutions for $\mbtheta$ or a closed-form expression of the divergence are unavailable, we may use the reparameterization trick to optimize $\mbtheta$. We demonstrate this approach using the maximum mean discrepancy in \ref{sec:appendix-other-divergences}.

\subsection{Bi-level optimization}\label{sec:alg3}
The inference process of FedPP is formulated as a bi-level optimization problem. The outer optimization aims to minimize the global objective function $F(\mbtheta)$, which is defined as the average of local objective functions across all clients:
\begin{equation}
    \min_{\mbtheta} F(\mbtheta) := \frac{1}{C}\sum_{c=1}^{C} F_c(\mbtheta), \label{eq:server-objective-function} 
\end{equation}
where $F_c(\mbtheta)$ is the local objective function for client $c$, defined as:
\begin{multline} \label{eq:client-objective-function}
    F_c(\mbtheta)  := \min_{\mbphi_c, \nu_c, m_c} \Big\{ \DV{q_{\mbphi_c}(\mbw_c)}{p_{\mbtheta}(\mbw_c)} 
     - \mathbb{E}_{q_{\mbphi_c}(\mbw_c)} \left[ \log p({\mbt}_c|\mbw_c, \mbz, m_c, \nu_c) \right] \Big\}
\end{multline}
In Eq.~\eqref{eq:server-objective-function}, $F(\mbtheta)$ represents the global objective function, and the local objective function $F_c(\mbtheta)$ in Eq.~\eqref{eq:client-objective-function} is minimized at the client level.

{For each client, we use a negative evidence lower bound (ELBO)-based objective function that consists of two main components: a reconstruction loss and a divergence term. The reconstruction loss is:
\begin{equation}
    -\mathbb{E}_{q_{\mbphi_c}(\mbw_c)} \left[ \log p({\mbt}_c|\mbw_c, \mbz, m_c, \nu_c) \right],
\end{equation}
and the divergence term is:
\begin{equation}
    \DV{q_{\mbphi_c}(\mbw_c)}{p_{\mbtheta}(\mbw_c)}.
\end{equation}
These two terms are jointly optimized to update the variational parameters $\mbphi_c$. Note that only the divergence term in Eq.~\eqref{eq:server-objective-function} depends on the global parameter $\mbtheta$.}

{In FedPP, the variational distribution of $\mbw_c$ participates in the communication between the server and clients, as learning $\mbw$ serves as a universal feature extractor across all clients. Meanwhile, the mean functions $\{\nu_c\}_{c=1}^C$ and the inducing points $\{\mbu_c\}_{c=1}^C$ of GP, and the scaling parameters $\{m_c\}_{c=1}^C$ of SGCP are optimized independently for each client, reflecting the unique characteristics and tasks of each client. This allows FedPP to act as a personalized federated learning framework, enabling client-specific models while maintaining global knowledge sharing.}

\subsection{The FedPP Algorithm}\label{sec:alg4}
{We propose a novel federated learning (FL) algorithm to instantiate FedPP in a practical cross-device FL setting, where a random subset $S_j$ of all clients participates in each communication round. Directly optimizing Eq.~\eqref{eq:client-objective-function} is challenging due to the absence of an analytical form for the reconstruction loss. To address this issue, we outline a three-step strategy.}

{\noindent\textbf{Step 1: Calculating ${\E}_{q_{\mbphi_c}({\mbw}_c)}[\log p({\mbt}_c|{\mbw}_c, {\mbz}, m_c, \nu_c)]$.}
The likelihood of events for client $c$, $p({\mbt}_c|\mbw_c, \mbz, m_c, \nu_c)$, is computed by integrating out the approximated function $\tilde{f}_c(t)$ and the inducing points $\mbu_c$:
\begin{equation}
\begin{aligned}
p({\mbt}_c|\mbw_c, \mbz, m_c, \nu_c) = \int p({\mbt}_c|m_c, \tilde{f}_c(t)) p(\tilde{f}_c(t)|\mbu_c, \mbw_c, \mbz, \nu_c) 
 p(\mbu_c|\mbz, \mbw_c) \, d\mbu_c \, d\tilde{f}_c(t),
\end{aligned}
\end{equation}
where $\tilde{f}_c(t)$ is the approximated function and $\mbu_c$ are the inducing points.

The expression $p({\mbt}_c|\mbw_c, \mbz, m_c, \nu_c)$ involves integration and sigmoidal operations in the exponential term, which makes direct inference challenging. To overcome this, we adopt the SGCPs model~\cite{donner2018efficient,zhou2019scalable,zhou2020efficient,xuhui2021neurips,zhou2023heterogeneous} and augment it with latent marked Poisson processes $\Pi_c$, where $m_c \times p_{\text{PG}}(\xi|1,0)$ serves as the intensity function. Additionally, P\'{o}lya-Gamma random variables $\{\xi_{c,i}\}_{i=1}^{n_c}$, governed by the P\'{o}lya-Gamma distribution $p_{\text{PG}}(\xi|1, 0)$, are incorporated. This allows us to express the augmented likelihood as:
\begin{multline}
    p({\mbt}_c|\tilde{f}_c(\cdot), m_c)  
    =\int p({\mbt}_c, \Pi_c, \{\xi_{c,i}\}_{i=1}^{n_c} |m_c, \tilde{f}_c(\cdot)) \\
    \cdot\prod_{(t_{c,j}, \xi_{c,j})\sim\Pi_c}\left[e^{g\left(\xi_{c,j}, -f_c(t_{c,j})\right)}m_c p_{\text{PG}}(\xi_{c,j}|1, 0)\right]
    \cdot d\Pi_c d\{\xi_{c,i}\}_{i=1}^{n_c},
\end{multline}
where $g(\xi, f)=f/2-\xi f^2/2-\log 2$. This results in a closed-form expression for the evidence lower bound (ELBO).}

{\noindent\textbf{Step 2: Mean-field Variational Inference for Random Variables $\mbu_c$, $\mbxi_c$, and $\mbPi_c$.} 
To simplify the inference, we assume independence between the random variables $\mbu_c$, $\mbxi_c$, and $\mbPi_c$. By using the standard mean-field variational inference method~\cite{bishop2006pattern,chen2023bayesian}, we compute the optimal variational distributions as follows:
\begin{align} \label{eq:genearal_MFVI_equation}
\ln q_{x_i}(x_i)={\E}_{q_{\backslash x_i}}\left[\ln p(\mathcal{D}, \{x_i\}_i)\right] + \text{const}.
\end{align}
It is noted that we are using KL-divergence between the variational distribution of $\mbu_c, \mbxi_c, \mbPi_c$ and their prior distribution here. The divergence $\DV{\cdot}{\cdot}$ is used for the case of $\mbw_c$ only. That is, we are using separate divergences for $\mbu_c, \mbxi_c, \mbPi_c$ and $\mbw_c$. 
Further details of the calculation for each of $\mbu_c, \mbxi_c, \mbPi_c$ are provided in \ref{app:details-of-mfvi}.}

{\noindent\textbf{Step 3: Maximizing the Local Objective Function w.r.t. $q_{\mbphi_c}({\mbw}_c)$.} The local objective function for client $c$ is abstracted as:
\begin{multline} \label{eq:phi-related-objective-function}
   L_c(\mbphi_c):=-{\E}_{q_{\mbphi_c}({\mbw}_c)}\left[\DV{q(\mbu_c)}{p(\mbu_c|\mbw_c, 
    {\mbz})}\right] \\   
    +{{\E}_{q_{\mbphi_c}({\mbw}_c)}\left[\log p(\mbt_c|\mbw_c, 
    -)\right]}-\DV{q_{\mbphi_c}({\mbw}_c)}{p_{\mbtheta}({\mbw}_c)}.
\end{multline}
We can leverage the reparameterization trick to compute a closed-form expression for the first two terms. The last term is straightforward to calculate since both $q_{\mbphi_c}(\mbw_c)$ and $p_{\mbtheta}(\mbw_c)$ follow Gaussian distributions.}

{\noindent\textbf{Brief Summary of the Local Updates:} At the $j$-th communication round, the global distribution $p_{\mbtheta^{(j)}}({\mbw}_c)$ is sent to each client $c$. During each local epoch, the variational distributions of $\mbu_c$, $\mbxi_c$, and $\mbPi_c$ are updated using Eq.~\eqref{eq:genearal_MFVI_equation}. Then, the variational parameters $\mbphi_c$ are optimized using first-order stochastic gradient descent (SGD):
\begin{equation} \label{eq:local-update-phi-c}
\mbphi_c^{(e)} \gets \mathrm{SGD}(-L_c(\mbphi_c); B, E, \eta),
\end{equation}
where $B$, $E$, and $\eta$ represent the mini-batch size, the number of local epochs, and the learning rate, respectively.}

\section{Related Work}
Since our work is the first to apply FL to predict events, we have a brief review on the FL with GPs and other typical methods in point process methods. 
\cite{DKL} explored GPs for few shot classification, which learns covariance functions parameterized by deep neural networks. \cite{VB-IB} applied GPs to meta-learning by maximizing the mutual information between the query set and a latent representation of the support set. 
\cite{yu2022federated} considered the GP FL by integrating deep kernel learning and scalable random features. \cite{achituve2021personalized} optimized the hyperparameters of a global deep kernel while training local GPs. 
DenseGP~\cite{denseGP} uses inducing locations in GP for meta learning, while using the same kernel parameters for all the clients/domains. {pFedBayes \cite{zhang2022personalized} may be closest to FedPP, which proposes a general framework of using Bayesian variational inference for FL. 
In addition to the application difference, pFedBayes uses an averaging strategy for global aggregation, which might be sub-optimal for the objective functions.} 

The TPPs with a nonlinear intensity function is important as it can model complex real-world patterns. These nonlinear functions include rectifier~\cite{reynaud2013inference}, exponential~\cite{gerhard2017stability} and sigmoid~\cite{linderman2016bayesian,apostolopoulou2019mutually}. The sigmoid mapping function has the advantage that the P\'{o}lya-Gamma augmentation scheme can be utilized to convert the likelihood into a Gaussian form, which makes the inference tractable. Also, there is a rich line of research developing neural embedding methods~\cite{du2016recurrent,mei2017neural,zhang2020self,zuo2020transformer,shchur2021neural,lin2022exploring} to study event data. Their main idea is to use various neural network architectures, including RNN, LSTM, and Transformer, to obtain historical embeddings, which differs substantially from our focus in this work. 

\section{Experiments}
\subsection{Data, Models and Settings} 
{In the experiments, we utilize five benchmark datasets from diverse domains: \textit{Taobao}, which contains timestamped user online shopping behaviors on the Taobao platform; \textit{Retweet}, consisting of timestamped user retweet events; \textit{Conttime}, a simulated public dataset provided by \cite{mei2017neural}; \textit{Stack Overflow}, which includes question-answering badge events; and \textit{Amazon}, featuring timestamped user purchase behaviors on the Amazon platform. These datasets are employed to evaluate the performance of our proposed model, FedPP, against baseline methods. For each sequence, we use the first $60\%$ for training, the next $20\%$ for validation (used for hyperparameter tuning), and the final $20\%$ for testing. Additionally, we normalize the entire observation timeline into $[0, 100]$ for numerical stability. To construct a heterogeneous setting, where local data across clients are non-IID, we allocate an equal number of samples to all clients, with each client being assigned $k (< K)$ different event types. More experimental details are provided in \ref{appendix:exp-details}.}

{\noindent\textbf{Comparison Methods.} FedPP is compared to the following baseline methods:
(1) \textit{RMTPPs} \cite{du2016recurrent}, which encodes past events using an RNN and employs a mixture of \textit{Gompertz} distributions for the probabilistic decoder;
(2) \textit{SAHP} \cite{zhang2020self}, which models the intensity function with a self-attention mechanism and uses the \textit{softplus} nonlinear transformation to ensure non-negativity of the intensity function;
(3) \textit{THP} \cite{zuo2020transformer}, a self-attention-based point process model similar to SAHP;
(4) \textit{TCVAE} \cite{lin2022exploring}, a generative neural embedding method that utilizes a variational autoencoder (VAE) for probabilistic decoding;
(5) \textit{TCDDM} \cite{lin2022exploring}, which is similar to TCVAE but uses a diffusion model for probabilistic decoding;
(6) \textit{Dec-ODE} \cite{song2024decoupled}, which employs neural ordinary differential equations (ODE) to capture complex dynamics in the intensity function;
(7) \textit{LSTM-FedPP}, the FedPP model with an LSTM-based historical encoder;
(8) \textit{ATT-FedPP}, the FedPP model with an attention-based historical encoder. As stated earlier, FedPP with deep kernel learning is the primary model under evaluation, while LSTM-FedPP and ATT-FedPP are left for future work.}

To facilitate fair and universally applicable comparisons, we implement federated modeling by applying parameter averaging to all the aforementioned original TPPs baseline methods, and prefixing them with ``Fed" to indicate their federated versions.

\textbf{Implementation.}
In all experiments, we set the total number of clients $C=20$. For each communication round, we fix the number of (randomly selected) participating clients to be $S=10$.
All datasets are configured with a fixed number of 100 global communication rounds, which is based on the observation that further increasing the number of communication rounds does not result in a significant performance improvement. We employ Adam \cite{kingma2014adam} as a basic optimizer and set the number of local epochs $E=5$ for all methods.

\noindent\textbf{Predictive Performance.} We adopt the expected posterior predictive log-likelihood on the test time interval to compare all the models. In order to sufficiently validate the modeling advantages of FedPP, we make different settings to evaluate its predictive performance. 

\subsection{Performance Comparison}
\textbf{Test Log-likelihood for Different Datasets.} We conduct the experiments under heterogeneous settings on five different benchmark datasets. Note that, since there are no closed-form likelihoods for Fed-TCVAE and Fed-TCDDM, we report the variational lower bounds\footnote{The performance of the closed-form likelihood is superior to that of the reported variational lower bound.} for them. The main results in Table~\ref{tab: comparison} indicate that our proposed method FedPP and its variants \texttt{LSTM-FedPP} and \texttt{ATT-FedPP} dominate the other baselines on all datasets except Conttime, which demonstrates the feasibility and effectiveness of our local model framework and the global aggregation scheme, especially for real-world datasets. Additionally, we observe that incorporating neural historical encoders consistently improves the model performance across all datasets. The performance of different neural historical encoders is similar, which aligns with the conclusion in \cite{lin2021empirical}.

\begin{table*}[t]
\renewcommand{\arraystretch}{1.0}
\caption{Performance evaluation for test log-likelihood (mean$\pm$std) on different benchmark datasets (\textbf{Bold} represents the top-3 performance).
  }
\label{tab: comparison}
\centering
\resizebox{\linewidth}{!}{
\begin{tabular}{ccccccc}
\toprule
 Dataset  &Taobao & Retweet & Conttime & Stack Overflow & Amazon\\
 \midrule
Fed-RMTPPs      & $1.34\pm0.12$ & $-1.73\pm0.06$ & $1.57\pm0.08$ & $-1.37\pm0.02$ & $1.24\pm0.13$   \\ 
Fed-SAHP       & $2.79\pm0.02$ & $\boldsymbol{-1.14\pm0.12}$  & $\boldsymbol{2.45\pm0.03}$ & $2.63\pm0.08$ & $3.92\pm0.04$  \\
Fed-THP        & $2.75\pm0.07$ & $-1.93\pm0.02$ & $2.34\pm0.07$  & $2.61\pm0.09$ & $3.96\pm0.02$ \\
Fed-TCVAE  & $\geq 1.26\pm0.06$ & $\geq -3.14\pm0.23$ & $\geq 1.28\pm0.09$ & $\geq 1.36\pm0.12$ & $\geq 1.97\pm0.11$   \\ 
Fed-TCDDM  & $\geq 1.22\pm0.12$ & $\geq -2.13\pm0.18$ & $\geq 1.34\pm0.08$  & $\geq 1.42\pm0.07$ & $\geq 2.38\pm0.06$ \\
Fed-Dec-ODE      & $2.88\pm0.01$ & $-1.35\pm0.08$  & $\boldsymbol{2.53\pm0.04}$ & $2.63\pm0.02$ & $3.93\pm0.03$ \\
\midrule
FedPP  & $\boldsymbol{2.92\pm0.04}$ & $-1.32\pm0.16$ & $2.42\pm0.16$ & $\boldsymbol{2.66\pm0.01}$ & $\boldsymbol{4.07\pm0.01}$ \\
\texttt{LSTM-FedPP}  & $\boldsymbol{3.14\pm0.05}$ & $\boldsymbol{-1.27\pm0.13}$ & $2.44\pm0.12$ & $\boldsymbol{2.68\pm0.07}$ & $\boldsymbol{4.12\pm0.04}$ \\
\texttt{ATT-FedPP}  & $\boldsymbol{3.07\pm0.02}$ & $\boldsymbol{-1.23\pm0.09}$ & $\boldsymbol{2.46\pm0.08}$ & $\boldsymbol{2.71\pm0.07}$ & $\boldsymbol{4.09\pm0.03}$ \\
\bottomrule
\end{tabular}
}
\end{table*}

\textbf{Ablation Study.} FedPP takes two critical strategies to improve its performance: 1) For the server, we design the global aggregation guided by divergence such as KL divergence, rather than the simple federated averaging, for a more reasonable global model, especially for heterogeneous data scenarios. 2) For each client, we add the deep kernel to transform inputs into the points in the input space of GP for a scalable and powerful uncertainty representation. To investigate the individual efficacy of these two strategies, we perform an ablation study on all datasets and report the test log-likelihood in Table~\ref{tab: ablation study}. 
This result indicates that both strategies individually improve the model performance, and using both methods simultaneously yields the best results. This is because using a deep kernel helps model the input, while utilizing KL aggregation assists in capturing the relationship between local models and the global model from a distributional perspective.

\begin{table*}[t]
\renewcommand{\arraystretch}{1.0}
\caption{Ablation study for FedPP. KL and DK represent the aggregation by the KL divergence and the deep kernel, respectively. `None' implies that neither of two strategies has been used, while `Both' indicates that both strategies have been employed.}
\label{tab: ablation study}
\centering
\resizebox{\linewidth}{!}{
\begin{tabular}{ccccccc}
\toprule
 Dataset  &Taobao & Retweet & Conttime & Stack Overflow & Amazon\\
 \midrule
None & $2.75\pm0.02$ & $-1.40\pm0.06$ & $2.12\pm0.08$ & $2.54\pm0.02$ & $3.82\pm0.04$   \\ 
w/KL & $2.78\pm0.02$ & $-1.42\pm0.05$  & $2.15\pm0.12$ & $2.57\pm0.04$ & $3.84\pm0.01$  \\
w/DK & $2.82\pm0.05$ & $-1.38\pm0.03$ & $2.37\pm0.03$  & $2.59\pm0.06$ & $3.91\pm0.02$ \\
Both & $\boldsymbol{2.92\pm0.04}$ & $\boldsymbol{-1.32\pm0.16}$ & $\boldsymbol{2.42\pm0.16}$  & $\boldsymbol{2.66\pm0.01}$ & $\boldsymbol{4.07\pm0.01}$ \\
\bottomrule
\end{tabular}
}
\end{table*}

\textbf{Effects of Different Global Aggregation Schemes.} In addition to the KL divergence aggregation, the Wasserstein distance is also an alternative aggregation method. In this experiment, we investigate the impact of FedAvg, KL divergence aggregation, and Wasserstein distance aggregation on model performance across five different benchmark datasets. The results in Table~\ref{tab: aggregation scheme} indicate that the improvement in model performance by KL aggregation and Wasserstein distance aggregation is superior to FedAvg, and the KL aggregation scheme slightly outperforms the Wasserstein distance scheme on all datasets.

\begin{table*}[t]
\renewcommand{\arraystretch}{1.0}
\caption{Comparison of test log-likelihood with different global aggregations on benchmark datasets.}
\label{tab: aggregation scheme}
\centering
\resizebox{\linewidth}{!}{
\begin{tabular}{ccccccc}
\toprule
 Dataset  &Taobao & Retweet & Conttime & Stack Overflow & Amazon\\
 \midrule
FedAvg & $2.82\pm0.03$ & $-1.38\pm0.05$ & $2.37\pm0.09$ & $2.59\pm0.03$ & $3.91\pm0.07$   \\ 
Wasserstein & $2.91\pm0.08$ & $-1.35\pm0.06$  & $2.40\pm0.12$ & $2.61\pm0.09$ & $4.05\pm0.07$  \\
KL & $\boldsymbol{2.92\pm0.04}$ & $\boldsymbol{-1.32\pm0.16}$ & $\boldsymbol{2.42\pm0.16}$  & $\boldsymbol{2.66\pm0.01}$ & $\boldsymbol{4.07\pm0.01}$ \\
\bottomrule
\end{tabular}
}
\end{table*}

\subsection{Further Evaluation on Federated Settings}
Since this work primarily explores the feasibility and effectiveness of the TPPs in federated scenarios, it is crucial to investigate the performance of the proposed method FedPP in various federated settings.

\subsubsection{Main Results on Local Epochs}
In general, there is a trade-off between local training steps and global communication rounds in FL, which means that adding more computation to each client in each communication round may reduce the overall communication cost of the FL system. 
In this experiment, we explore this characteristic by varying the values of the local epochs $E$. 
Specifically, we conduct the experiment on several benchmark datasets and set the number of local epochs between communication rounds over $\{1, 2, 5, 10\}$ with the total communication rounds fixed. The results are shown in Fig. \ref{fig:3-results} (left).  
Except for Stack Overflow, more local epochs lead to worse performance. 
Particularly for Conttime, more local epochs result in obvious performance degradation. However, it is worth noting that a balance between computation and communication costs is essential in selecting local epochs. 

\subsubsection{Main Results on Highly Data-scarce Cases}
\begin{figure*}[t]
\centering
\includegraphics[width = 0.4\textwidth]{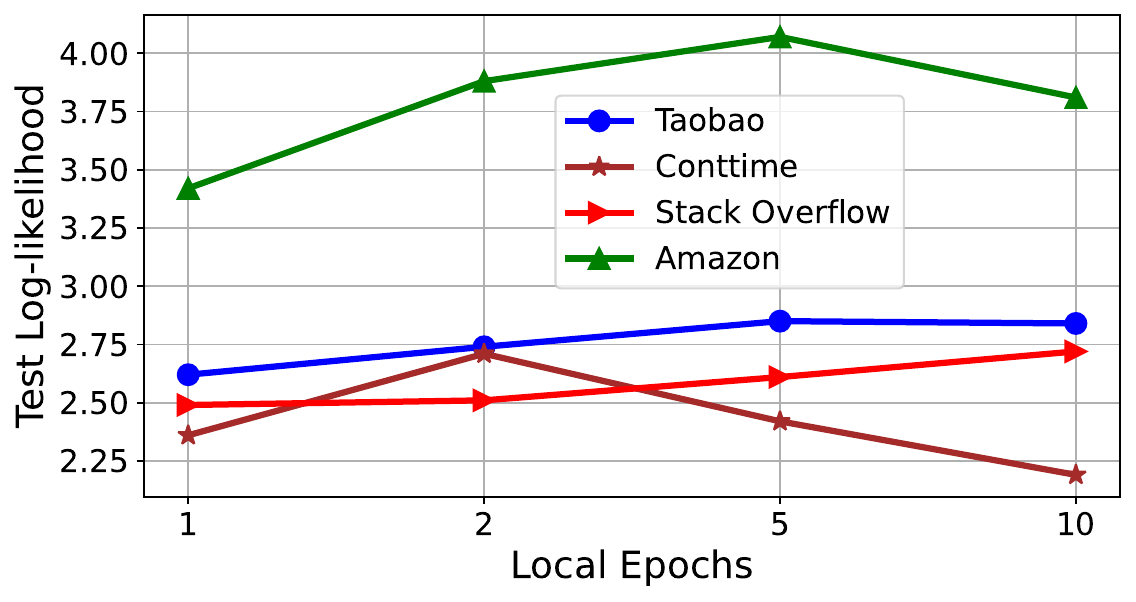}
\includegraphics[width =0.28\textwidth]{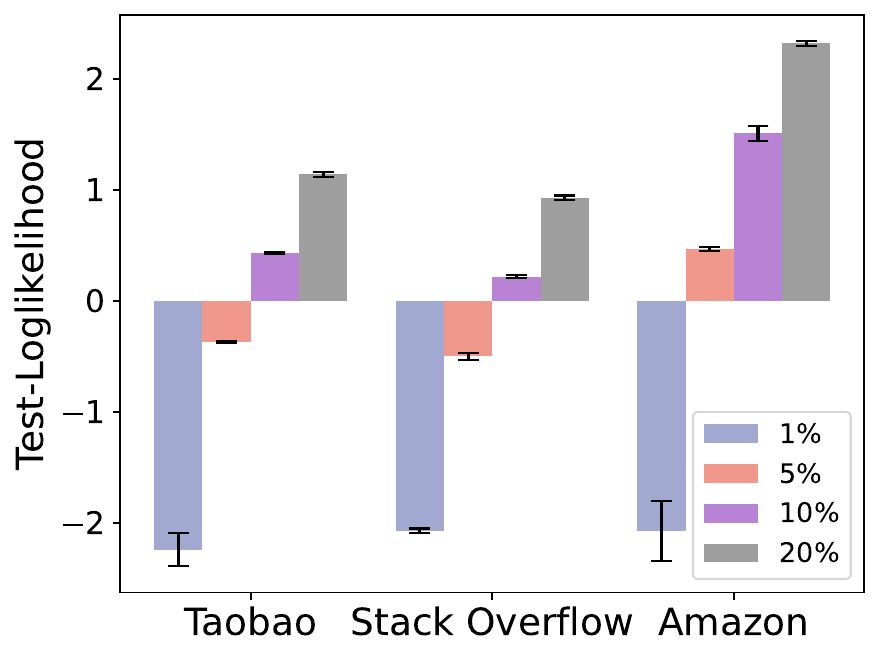}
\includegraphics[width =0.28\textwidth]{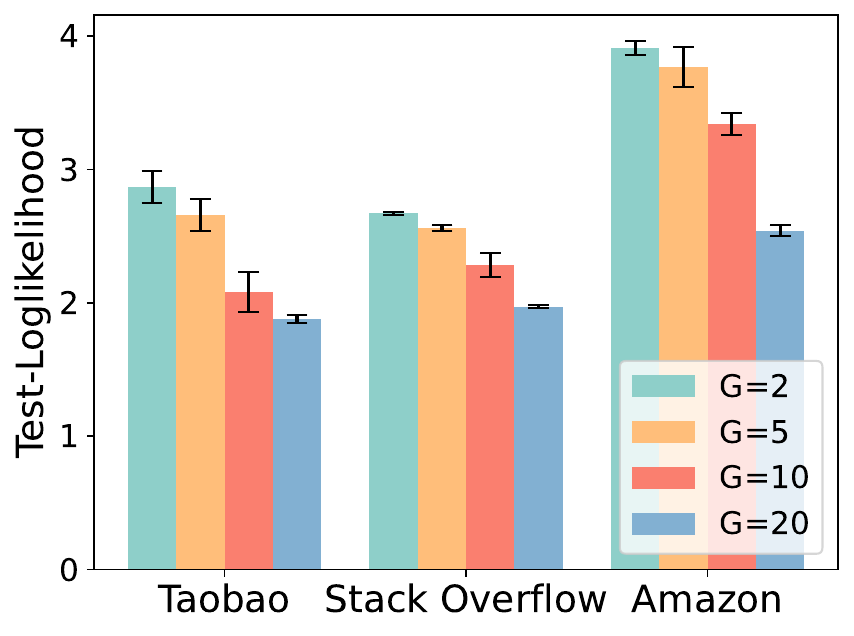}
\caption{(Left) Performance comparison of test log-likelihood on four different datasets with varying local epochs. (Middle) Performance comparison of test log-likelihood on three different datasets by varying the local data size. (Right) Performance comparison w.r.t. test log-likelihood on three different datasets by varying the struggling period.}
\label{fig:3-results}
\vspace{-0.5cm}
\end{figure*}

To evaluate the impact of local data size on the model performance, we conduct the experiment on highly data-scarce cases. Based on the default experimental data size, we further vary the proportion of local data size for each client over $\{1\%, 5\%, 10\%, 20\%\}$ on Taobao, Stack Overflow, and Amazon. The results of different datasets are reported in Fig. \ref{fig:3-results} (middle). It can be found that the absence of data does make a certain impact on the model performance, especially in the case of the $1\%$ local data size, where the model performance experiences severe degradation. However, when the data size is increased by just $5\%$, our method obtains relatively good results, demonstrating the superior performance of our method in such extreme federated settings. 

\subsubsection{Main Results on Straggling Scenarios}
As we know, FL is a dynamic learning system, inactive or unparticipated clients in each communication round are random and unpredictable. A highly challenging scenario is when some clients experience a long straggling period, leading to a long-term absence of certain training samples \cite{xu2023stabilizing}. To explore the robustness of the proposed approach in such scenarios, we conduct the experiments on Taobao, Stack Overflow and Amazon  and set the straggling period $G$ to 2, 5, 10 and 20 communication rounds respectively for comparison. 
Namely, the activated clients and inactivated clients remain fixed for every $G$ communication rounds. Fig. \ref{fig:3-results} (right) presents the test log-likelihood over three different real-world datasets, from which we can see that larger straggling periods $G$ has a stronger negative impact on the model performance. This is mainly because some 
inactive clients have important features that do not present in currently participating clients, causing bias in the global model learned during the straggling period. Additionally, we observe that our method achieves relatively smooth results on these three datasets, especially on Stack Overflow. This demonstrates the stability and reliability of our method in such FL scenarios.

\section{Conclusions}
We have introduced FedPP, the first approach to learning point processes for uncertain and sparse events in a federated and privacy-preserving manner. Extensive experimental evaluations show that FedPP consistently outperforms federated versions of benchmark methods across five datasets. The integration of neural embeddings into Sigmoidal Gaussian Cox Processes proves highly effective in both summarizing historical information and learning flexible intensity functions. Additionally, we proposed a novel framework for global aggregation using divergence measures, offering practical interpretations for the aggregation process. This divergence-guided framework can be extended to any Bayesian federated learning method that involves probability distribution aggregation, paving the way for future work in this direction.

\appendix
\section{Related work about Survival analysis}
Survival analysis is a statistical framework used to analyze the time until an event of interest occurs, commonly employed in fields such as healthcare, finance, and engineering to model time-to-event data \cite{jenkins2005survival,wang2019machine}. Traditional survival analysis often relies on techniques like the Cox proportional hazards model, where the risk (or hazard) of an event is estimated over time. In this context, events are usually assumed to occur only once, and the data might involve censoring, where the event does not occur during the observation period. When extended to FL, survival analysis must adapt to decentralized settings where data is distributed across multiple clients, such as hospitals or devices \cite{rahimian2022practical}, without sharing raw data for privacy reasons \cite{rahman2023fedpseudo,rahman2023communication}. In federated learning-based survival analysis, each client locally trains a survival model on its data, and the central server aggregates the learned models without accessing individual-level data, thus maintaining privacy.

In contrast, TPPs focus on predicting multiple recurring events over time, such as transaction patterns, server requests, or patient visits. Unlike survival analysis, which typically handles a single event per subject and often deals with censoring, TPPs model event intensity functions to capture sequential occurrences and inter-event dependencies. When extended to federated learning, TPP-based models face similar privacy-preserving challenges, but with the added complexity of modeling dynamic temporal relationships between multiple events. This distinction makes TPP-based federated models more flexible in applications where multiple events are expected, but it also demands more sophisticated techniques to capture the intricate dependencies and ensure global model aggregation across clients. Hence, while both survival analysis and TPP share common goals in event prediction, the nature of events and the modeling approaches differ significantly in their federated learning extensions.

\section{Deep Kernel Learning} 
Modern GP approaches, such as Deep Gaussian Processes~(DGPs)~\cite{damianou2013deep} and Deep Kernel Learning~(DKL)~\cite{wilson2016deep}, usually use deep architectures to enhance the expressive power of kernels. For instance, DKL defines kernels on features mapped by a neural network $g_{{\tilde{\mbw}}}(\cdot)$, with ${\tilde{\mbw}}$ being the neural network parameters. In this way, the RBF kernel entry between $\mbx_i$ and $\mbx_j$, as an example, can be formulated as $\kappa_{{\mbx}_i,{\mbx}_j;[\tilde{\mbw};r;l]}=r\cdot\exp(-{\|g_{\tilde{\mbw}}({\mbx}_i)-g_{\tilde{\mbw}}({\mbx}_j)\|^2}/({2l^2}))$, with $r$ and $l$ being the scaling and band-width parameters. Without particularly specified, we use the RBF kernel as the default kernel in this work.

\section{Additivity Property of TPPs}
\begin{property} \label{property:additivity-of-TPPs}
    Let $N_1(t)$ and $N_2(t)$ be two independent temporal point processes with intensity functions $\lambda_1(t)$ and $\lambda_2(t)$ respectively. The process $N(t) = N_1(t) + N_2(t)$ is also a temporal point process with the intensity function $\lambda(t) = \lambda_1(t) + \lambda_2(t)$. 
\end{property}

\section{Sparse Gaussian Processes} \label{sec:sparse-gp}
Although powerful and highly flexible, GP, which is the key component in SGCPs, is often questioned due to its high computational cost. 
In particular, obtaining the posterior of the latent function requires inverting the kernel matrix $\kappa_{{\mbx}_{1:N}, {\mbx}_{1:N};{\mbw}}$ of $N$ training data points ${\mbx}_{1:N}$, resulting in a computational cost scaled to $\mathcal{O}(N^3)$. 
Introducing sparse inducing points ${\mbu}$ is a common approximation strategy to mitigate this issue. Specifically, the kernel matrix $\kappa_{{\mbz}, {\mbz};{\mbw}}$ over $M$ inducing inputs ${\mbz}_{1:M}$ can be regarded as a low-rank approximation to the full kernel matrix $\kappa_{{\mbx}_{1:N}, {\mbx}_{1:N};{\mbw}}$, reducing the computational cost from $\mathcal{O}(N^3)$ to $\mathcal{O}(NM^2)$. Given inducing inputs ${\mbz}$ and inducing points ${\mbu}$, in which ${\mbu}$ is generated as ${\mbu}\sim\cN\left(\nu({\mbz}), \kappa_{{\mbz}, {\mbz};\mbw}\right)$, we can have the approximated random function value $\tilde{f}(\cdot)$ following the Gaussian distribution as: 
\begin{align} \label{eq:inducing_points_on_new_points}
\tilde{f}(\cdot)|{\mbw}, {\mbu}, {\mbz}\sim\cN(\tilde{f}(\cdot);\nu(\cdot)+\kappa_{\cdot,{\mbz};\mbw}\kappa_{{\mbz},{\mbz};\mbw}^{-1}({\mbu} \nonumber\\
-\nu({\mbz})), 
\kappa_{\cdot,\cdot;{\mbw}}-\kappa_{\cdot,{\mbz};\mbw}\kappa_{{\mbz},{\mbz};\mbw}^{-1}\kappa_{{\mbz},\cdot;\mbw}). 
\end{align}

\section{Steps of the Inference} \label{sec:step-1-calculation}
\noindent\textbf{Step 1, Calculating ${\E}_{q_{\mbphi_c}({\mbw}_c)}[\log p({\mbt}_c|{\mbw}_c, {\mbz}, m_c)]$}.
Since we model the intensity function of  client $c$ as $\lambda_c(t)=m_c\cdot\sigma(f_c(t))$, the likelihood of the event ${\mbt}_c$ can be approximated as:
{ \begin{align} \label{eq:likelihood_sgcp}
    p({\mbt}_c|{\mbw}_c, {\mbz}, m_c)
    \overset{(i)}{\approx}\int p({\mbt}_c|\tilde{f}_c(\cdot), m_c)p(\tilde{f}_c(\cdot)|{{\mbu}_c}, {\mbw}_c, {\mbz}) \nonumber\\
    p({{\mbu}_c}|{\mbw}_c,{\mbz})d{{\mbu}_c}d\tilde{f}_c(\cdot)
\end{align}}
in which $p({\mbt}_c|\tilde{f}_c(\cdot), m_c)$ can be written as:
\begin{align} \label{eq:likelihood_f_m}
    p({\mbt}_c|\tilde{f}_c(\cdot), m_c)
= \exp{(-\int_{0}^Tm_c\sigma(f_c(t))dt)}\prod_{i=1}^{n_c}m_c\sigma(f_c(t_{c,i})).
\end{align}

In Eq.~\eqref{eq:likelihood_sgcp}, we first augment the original likelihood $p({\mbt}_c|{\mbw}, {\mbz}, m_c)$ with random function $\tilde{f}_c(\cdot)$ and inducing points ${{\mbu}_c}$. The approximation $(i)$ in Eq.~\ref{eq:likelihood_sgcp} denotes that the distribution of random function $p(f_c)$ is approximated by $\tilde{P}(f_c)$, which is supported by the sparse inducing points ${{\mbu}_c}$~(see details in Eq.~\eqref{eq:inducing_points_on_new_points}). 

Eq.~\eqref{eq:likelihood_f_m} contains integration and sigmoidal operations in the exponential part, which makes it difficult to directly obtain a closed-form representation for inference schedules. 
Adapting the SGCPs methods~ \cite{donner2018efficient,zhou2019scalable,zhou2020efficient,xuhui2021neurips} and further augmenting latent marked Poisson processes $\Pi_c$, which has $\nu\times P_{\text{PG}}(\xi|1,0)$ as the intensity function in $[0,T]\times\mathbb{R}^+$, and Polya-Gamma random variables $\{\xi_{c,i}\}_{i=1}^{n_c}$, which has the prior distribution $P_{\text{PG}}(\xi|1, 0)$, into Eq.~\eqref{eq:likelihood_sgcp}, we have
{\begin{multline}
    p({\mbt}_c|\tilde{f}_c(\cdot), m_c)  
    =\int p({\mbt}_c, \Pi_c, \{\xi_{c,i}\}_{i=1}^{n_c} |m_c, \tilde{f}_c(\cdot))d\Pi_cd\{\xi_{c,i}\}_{i=1}^{n_c} \\
    =\int e^{-m_c T}\cdot\prod_{i=1}^{n_c}\left[m_c e^{g\left(\xi_{c,i}, f_c(t_{c,i})\right)}P_{\text{PG}}(\xi_{c,i}|1, 0)\right] \\
    \prod_{(t_{c,j}, \xi_{c,j})\sim\Pi_c}\left[e^{g\left(\xi_{c,j}, -f_c(t_{c,j})\right)}m_c P_{\text{PG}}(\xi_{c,j}|1, 0)\right]
    \cdot d\Pi_cd\{\xi_{c,i}\}_{i=1}^{n_c}
\end{multline}
}
in which $g(\xi, f)=f/2-\xi f^2/2-\log 2$. 

In short, through augmenting the random function $\tilde{f}_c(\cdot)$, inducing points ${{\mbu}_c}$, latent marked Poisson processes $\Pi_c$, and Polya-Gamma random variables $\{\xi_{c,i}\}_{i=1}^{n_c}$, we are able to obtain a closed-form representation of ELBO for client $c$, which is:
 \begin{align}
     &\hspace{-.5cm} {\E}_{q_{\mbphi_c}({\mbw}_c)}[\log p({\mbt}_c|{\mbw}, {\mbz}, m_c)] -\DV{q_{\mbphi_c}({\mbw}_c)}{p({\mbw}_c)} \nonumber\\
    \ge& {\E}_{q(\tilde{f}_c(\cdot))}\left[\log P\left({\mbt}_c\g m_c, \tilde{f}_c(\cdot)\right)\right]
     \nonumber\\
     &-\KL{q\left({\mbu}_c, \tilde{f}_c(\cdot)\right)}{P\left({\mbu}_c, \tilde{f}_c(\cdot)\right)}-\DV{q_{\mbphi_c}({\mbw}_c)}{p({\mbw}_c)} \nonumber\\
    \ge& {\E}_{q(\cdot)}\left[\log P\left({\mbt}_c|\Pi_c, \{\xi_{c,i}\}_{i=1}^{n_c}, m_c, \tilde{f}_c(\cdot)\right)\right]
     -\KL{q\left({\mbu}_c, \tilde{f}_c(\cdot)\right)}{P\left({\mbu}_c, \tilde{f}_c(\cdot)\right)}\nonumber\\
     &-\KL{q\left(\Pi_c, \{\xi_{c,i}\}_{i=1}^{n_c}\right)}{P\left(\Pi_c, \{\xi_{c,i}\}_{i=1}^{n_c}\right)}
    -\DV{q_{\mbphi_c}({\mbw}_c)}{p({\mbw}_c)} \nonumber\\
    &=  \ELBO_c .  \label{eq:clients-final-elbo}
\end{align}
In fact, we are able to obtain closed-form representations for all the terms involved in Eq.~\eqref{eq:clients-final-elbo} and then use mean-field variational inference to obtain closed-form updating equations for the variational distributions of these random variables. \\

\noindent\textbf{Step 3, Maximizing the Objective Function of Client $c$ with Respect to $q_{\mbphi_c}({\mbw}_c)$.} We use the first-order stochastic gradient descent (SGD) algorithms to optimize the ELBO with respect to $q_{\mbphi_c}({\mbw}_c)$. For  problem (5) of clients, we can get the closed-form results for KL divergence terms but cannot get that for the other term. To speedup the convergence, we use the minibatch gradient descent (GD) algorithm. So the stochastic estimator for the $c$-th client is given. Regarding $q_{\mbphi_c}(\mbw_c)$, its corresponding terms in $\ELBO_c$ can be abstracted as:
\begin{equation}
\begin{aligned}
\label{eq-elbo-c-supp}    
    &-{\E}_{q_{\mbphi_c}({\mbw}_c)}\left[\DV{q(\mbu_c)}{p(\mbu_c|\mbw_c, 
    {\mbz})}\right]\\
    &+{\E}_{q_{\mbphi_c}({\mbw}_c)}\left[\log p(f_c(\cdot)|\mbw_c, 
    {\mbz})\right]
    -\DV{q_{\mbphi_c}({\mbw}_c)}{p({\mbw}_c)}.
\end{aligned}
\end{equation} 
We set the variational distribution $q_{\mbphi_c}(\mbw_c)$ as
\begin{equation} 
q_{\mbphi_c}(\mbw_c)=\cN([\mbw;\mbz];[\mbr_{\theta};\mbr_{z}],\diag\left( [\mbdelta_{\theta}^2;\mbdelta_{z}^2])\right).
\end{equation} 
The KL divergence term can be obtained in an analytical format since both $q_c=(\mbw,\mbz)$ and $\pi_c=(\mbw,\mbz)$ are Gaussian distributions, which is:
\begin{equation}
     \DV{q_{\mbphi_c}({\mbw}_c)}{p({\mbw}_c)}
    \propto  \frac{1}{2}\sum_m\bigg(\frac{(\mbmu_{\theta,z,m}-\mbr_{\theta,z,m})^2}{\mbsigma_m^2} \nonumber\\
    +\frac{\delta^2_m}{\sigma^2_m}-\log\frac{\delta^2_m}{\sigma_m^2}\bigg).
\end{equation}
Regarding the first term in Eq.~\eqref{eq-elbo-c-supp}, we can use reparameterization trick to formulate it as: 
\begin{equation}
\begin{aligned}
    &{\E}_{q_{\mbphi_c}({\mbw}_c)}\left[\DV{q(\mbu_c)}{p(\mbu_c|\mbw_c,{\mbz})}\right] \\
=&    \frac{1}{2}{\E}_{q_{\mbphi_c}({\mbw}_c)}\bigg[\log\frac{|\kappa_{\mbz,\mbz;\mbw}|}{|\Sigma_s|}+\text{trace}(\kappa_{\mbz,\mbz;\mbw}^{-1}\Sigma_s)+(\mbmu_s-\mbnu)^{\top}\kappa_{\mbz,\mbz;\mbw}^{-1}(\mbmu_s-\mbnu)\bigg] \\
=&    \frac{1}{2}{\E}_{{\mbepsilon_{\theta},\mbepsilon_z\sim\cN(0,\mbI)}}\bigg[\log\frac{|\kappa_{\mbz,\mbz;\mbw}|}{|\Sigma_s|}+\text{trace}(\kappa_{\mbz,\mbz;\mbw}^{-1}\Sigma_s)+(\mbmu_s-\mbnu)^{\top}\kappa_{\mbz,\mbz;\mbw}^{-1}(\mbmu_s-\mbnu)\bigg]
\end{aligned}
\end{equation} 
Regarding the second term in Eq.~\eqref{eq-elbo-c-supp}, it can be formulated as:
\begin{multline}
    {\E}_{q_{\mbphi_c}({\mbw}_c)}\left[\log p(f_c(\cdot)|\mbw_c, 
    {\mbz})\right] \\
    \propto \int {\E}_q\left[-\frac{f(t)}{2}+\frac{f^2(t)}{2}\xi\right]\Lambda(t,\xi)dt+
    \sum_{n}{\E_q}\left[\frac{f(t_n)}{2}+\frac{f^2(t_n)}{2}\xi_n\right].
\end{multline}

\section{Details of Mean-field Variational Inference} \label{app:details-of-mfvi}
\textbf{Optimal Polya-Gamma Density $q(\xi_{c,m})$}. 
\begin{align}
\log q(\xi_{c,m})={\E}_{q}\left[-[f_c]^2(t_m)\xi_{c,m}/2\right]+\log P_{\text{PG}}(\xi_{c,m}|1,0)+\text{const}
\end{align}
Thus, we get
\begin{align}
q(\xi_{c,m})\propto\exp\left[-{\E}_q([f_c]^2(t_m))\xi_{c,m}/2\right]\cdot P_{\text{PG}}(\xi_{c,m}|1,0)
\end{align}
which leads to
\begin{align} \label{eq:polya_gamma_1}
q(\xi_{c,m})=P_{\text{PG}}\left(\xi_{c,m}|1,\sqrt{{\E}_q([f_c]^2(t_m))}\right)
\end{align}
We can let $c_{c,m}=\sqrt{{\E}_q([f_c]^2(t_m))}$.

\textbf{Optimal Poisson Process $q(\Pi_c)$}.
Using the mean-field updating mechanism, we get the rate functions for the latent marked Poisson processes as:
\begin{align} \label{eq:latent_marked_PP_1}
\Lambda_c(t, \xi)=\frac{\exp({\E}_q\left[\log {\nu}_c\right]-\frac{{\E}_q\left[f_c(t)\right]}{2})}{2\cosh(\frac{c_c(t)}{2})}P_{\text{PG}}(\xi|1, c_c(t))
\end{align}
where $c_c(t)=\sqrt{{\E}_q[f_c(t)]^2]}$. Again, we emphasize that the support of $\Lambda_c(t, \xi)$ is $(0, T]\times \mathbb{R}^+$.


\textbf{Optimal Gaussian Processes $f_c(t)$}.
For the notation convenience, we neglect  script $c$ in the following discussion. 
\begin{align}
\log q(f)\propto \exp(U(f))
\end{align}
$U(f)$ is defined as:
\begin{align}
U(f)&={\E}_{q}\left[\sum_{(x,\xi)_n\in\Pi_{}}h(\xi_{n}, -f(x_{n}))\right]+\sum_{m}{\E}_q\left[h(\xi_{m},f(t_m))\right]\\
&=  -\frac{1}{2}\int_{\mathcal{T}}A(t)f(t)dt +\int_{\mathcal{T}}B(x)f^2(t)dt
\end{align}
where 
\begin{align}\label{eq:gp_f_1}
A(t)=\sum_{m}{\E}_q[\xi_{m}]\delta(t-t_m)+\int_{0}^{\infty}\xi\Lambda_{}(t,\xi)d\xi 
\end{align}
\begin{align}\label{eq:gp_f_2}
B(t)= \frac{1}{2}\sum_{m}\delta(t-t_m)-\frac{1}{2}\int_{0}^{\infty}\Lambda_{}(t,\xi)d\xi
\end{align}
in which we have the following equations for the above integration:
\begin{align}\label{eq:gp_f_3}
{\E}_q[\xi_{m}]=\frac{1}{2c_{m}}\tanh\left(\frac{c_{m}}{2}\right)
\end{align}
\begin{align}\label{eq:gp_f_4}
\int_{\mathcal{T}}\Lambda_{}(t,\xi)d\xi=\Lambda_{}(t)
\end{align}
\begin{align}\label{eq:gp_f_5}
\int_{\mathcal{T}}\xi\Lambda_{}(t, \xi)d\xi=  \frac{1}{2c_{}(t)}\tanh\left(\frac{c_{}(t)}{2}\right)\Lambda_{}(t)
\end{align}

By using sparse Gaussian process methods and its extension to infinite dimensional problems~\cite{batz2018approximate,donner2018efficienta}, we get the sparse posterior distribution for the function values at the inducing points as:
\begin{align} \label{eq:gp_f_6}
q(\mbu)=\cN(\mbmu_s, \Sigma_s)
\end{align}
where the covariance matrix and mean are expressed as:
\begin{align}\label{eq:gp_f_7}
\Sigma_s =  \left[\kappa_{\mbz,\mbz;\mbw}^{-1}\int_{\mathcal{T}}A(t)\kappa_{\mbz,t;\mbw}\kappa_{t,\mbz;\mbw}dt\kappa_{\mbz,\mbz;\mbw}^{-1}+\kappa_{\mbz,\mbz;\mbw}^{-1}\right]^{-1}
\end{align}
\begin{align} \label{eq:gp_f_8}
\mbmu_s=  \Sigma_s\left(\kappa_{\mbz,\mbz;\mbw}^{-1}\int_{\mathcal{T}}\tilde{B}(t)\kappa_{\mbz,t;\mbw}dt+\kappa_{\mbz,\mbz;\mbw}^{-1}\nu\right)
\end{align}
where $\tilde{B}(t)=B(t)-A(t)(\nu-\kappa_{t,\mbz;\mbw}\kappa_{\mbz,\mbz;\mbw}^{-1}\mbnu)$.

\section{Divergence} \label{sec:appendix-other-divergences}
%
%
The proofs of the following propositions follow immediately.
\begin{proposition}
    Given $x_1\sim\cN(x_1;\mu_1,\sigma_1^2), x_2\sim\cN(x_2;\mu_2,\sigma_2^2)$ and for any $\delta>0$, we can obtain the following:
    \begin{align}
        \mathbb{E}_{x_1\sim\cN(x_1;\mu_1,\sigma_1^2), x_2\sim\cN(x_2;\mu_2,\sigma_2^2)}\left[e^{-\frac{(x_1-x_2)^2}{\delta^2}}\right]=\delta\cdot\frac{e^{-\frac{(\mu_1-\mu_2)^2}{\delta^2+2\sigma_1^2+2\sigma_2^2}}}{\sqrt{\delta^2+2\sigma_1^2+2\sigma_2^2}}
    \end{align}
\end{proposition}

\begin{proposition}
    Given $\cN(\mu_1,\sigma_1^2), \cN(\mu_2,\sigma_2^2)$ and for any $\delta>0$, the maximum mean discrepancy (MMD) of the RBF kernel can be expressed as:
    \begin{align}
        &\text{MMD}(\cN(\mu_1,\sigma_1^2)\| \cN(\mu_2,\sigma_2^2))\\
        =&\delta\cdot\frac{1}{\sqrt{\delta^2+4\sigma_1^2}}+\delta\cdot\frac{1}{\sqrt{\delta^2+4\sigma_2^2}}-2\delta\cdot\frac{e^{-\frac{(\mu_1-\mu_2)^2}{\delta^2+2\sigma_1^2+2\sigma_2^2}}}{\sqrt{\delta^2+2\sigma_1^2+2\sigma_2^2}}
    \end{align}
\end{proposition}
\begin{proposition}
    $\sum_{c=1}^C MMD[{q_{\mbphi_c}({\mbw}_c)\|p_{\mbtheta}({\mbw}_c)]}$ reaches its minimum for $\mbtheta=\{\mbmu,\mbsigma^2\}$ when $\mbtheta$ is valued at:
    \begin{align} \label{eq:wasserstein-distance}
\mbmu, \mbsigma^2:=\argmin_{\mbmu, \mbsigma^2}\sum_{c=1}^C\frac{1}{\sqrt{\delta^2+4\mbsigma^2}}-2\frac{e^{-\frac{(\mbr_c-\mu)^2}{\delta^2+2\delta_c^2+2\mbsigma^2}}}{\sqrt{\delta^2+2\delta_c^2+2\mbsigma^2}}
    \end{align}
\end{proposition}



\section{Algorithm}
 Algorithm \ref{alg1} shows the training procedure of FedPP. 
\begin{algorithm}
	\renewcommand{\algorithmicrequire}{\textbf{Server executes:}}
	\renewcommand{\algorithmicensure}{\textbf{Client} Update($c, \mbtheta^{(j)}$)$\boldsymbol{:}$}
	\caption{FedPP: Federated Point Process Algorithm}
	\label{alg1}
	\begin{algorithmic}
	\STATE \textbf{Input:} $J$-communication rounds, $E$-local epochs, $\mathbb{S}$-random subset of all clients. Initialize: $\mbmu^0$ and $\mbsigma^0$.
       \REQUIRE
	\FOR{$j=0,1,\dots, J-1$}
        \STATE Sample $\mathbb{S}^j$ clients with size $S$ uniformly at random
        \FOR{each client $c \in \mathbb{S}^j$ \textbf{in parallel}}
        \STATE $\mbphi^{j+1}_{c} \gets$ Client Update($c, \mbtheta^{(j)}$)
        \ENDFOR
        \STATE $\mbtheta^{(j+1)}=\argmin_{\mbtheta}\sum_{c\in S^j}\DV{q_{\mbphi_c(\omega_c)}}{p_{\mbtheta}(\omega_c)}$
        \ENDFOR
        \ENSURE
        \FOR{$e=1,2, \dots, E$}
        \STATE Update variational distributions of $\mbu_c$, $\mbxi_c$ and $\mbPi_c$ using Eq.~\eqref{eq:genearal_MFVI_equation}
        \STATE Update $\mbphi_c$ using Eq.~\eqref{eq:local-update-phi-c}
        \ENDFOR   
        
        \textbf{Return} $\mbphi^{(j)}_{c}: = \mbphi^{(E)}$
	\end{algorithmic}  
\end{algorithm}

\section{Experimental Details}\label{appendix:exp-details}

\textbf{Benchmark Datasets.}
We choose five benckmark datasets to evaluate our method: (1) \textit{Taobao}, which is a public dataset generated for the 2018 Tianchi Big Data Competition. It comprises timestamped behavioral records (such as browsing and purchasing activities) of anonymized users on the online shopping platform Taobao. The data spans from November 25 to December 03, 2017. The dataset comprises a total of 5,318 sequences and $K=17$ types of events. (2) \textit{Retweet}, which contains a total of 24,000 retweet sequences, where each sequence is composed of events represented as tuples, indicating tweet types and their corresponding times. There are $K=3$ distinct types of retweeters: small, medium, and large. To classify retweeters into these categories, small retweeters have fewer than 120 followers, medium retweeters possess more than 120 but fewer than 1,363 followers, and the remaining retweeters are labeled as large. The dataset provides information on when a post will be retweeted and by which type of user. (3) \textit{Conttime}, which is a public dataset releases by \citep{mei2017neural}. There are 9,000 sequences with a total of $K=5$ event types. (4) \textit{Stack Overflow}, which is a public dataset that encompasses sequences of user awards spanning a two-year period. In the Stack Overflow question-answering platform, recognizes users through awards based on their contributions, including posing insightful questions and providing valuable answers. The dataset comprises a total of 6,633 sequences and $K=22$ types of events. (5) \textit{Amazon}, which is a public dataset similar to Taobao, containing a timestamped behavioral record of anonymized users on the online shopping platform Amazon. There are 14,759 sequences with a total of $K=16$ event types.

Table~\ref{tab:experiments-datasets-statistics} displays the general characteristics of each dataset.

\begin{table}[t]
    \centering
    \caption{The statistics of five benchmark datasets}
    \label{tab:experiments-datasets-statistics}
    \begin{tabular}{cccccc}
    \toprule
  \multirow{2}{*}{Dataset}  & \multirow{2}{*}{$\#$ of sequences} & \multirow{2}{*}{Types-$K$} & \multicolumn{3}{c}{Sequence Length}  \\
   \cmidrule{4-6} & & &  Min & Mean & Max  \\
   \midrule
Taobao & 4800 & $17$ & $58$ & $59$ & $59$ \\
Retweet & 24000 & $3$ & $50$ & $109$ & $264$\\
Conttime & 9000 & $5$ & $20$ & $60$ & $100$\\
Stack Overflow & 6633 & $22$ & $41$ & $72$ & $736$\\
Amazon &  5200 & $16$ & $14$ & $45$ & $94$ \\
    \bottomrule
      \end{tabular}
\end{table}

\textbf{Training Details.} We set the number of clients $C$ to 20 and random subset of all clients $S$ to 10 for all approaches. In addition, we perform 5 epochs to train local model for each client. In heterogeneous setting, the number of event types for each client is set to $k=2$ for Retweet and Conttime, and $k=4$ for other datasets. For the deep kernel function, we use 1-layer Bayesian neural network as the deep learning module. For the historical encoder, we set the layer number and embedding size to 2 and 32, respectively. We tune the learning rate $\eta$ over \{1e-3, 5e-3, 1e-2, 5e-2, 0.1\} and fix the rate to 1e-3.
All experiments in this paper are implemented on a system comprising 4 cores, with each core powered by an Intel(R) Xeon(R) CPU E5-2686 at a frequency of 2.30GHz. The system is further enhanced by the inclusion of one NVIDIA Tesla K80 GPU. In addition to the GPU memory, the system boasts a total of 60GB of memory.

\section{Additional Experiments}
\subsection{Visualisation for the synthetic data}

\begin{figure}[t]
\centering
\includegraphics[width = 0.85\textwidth]{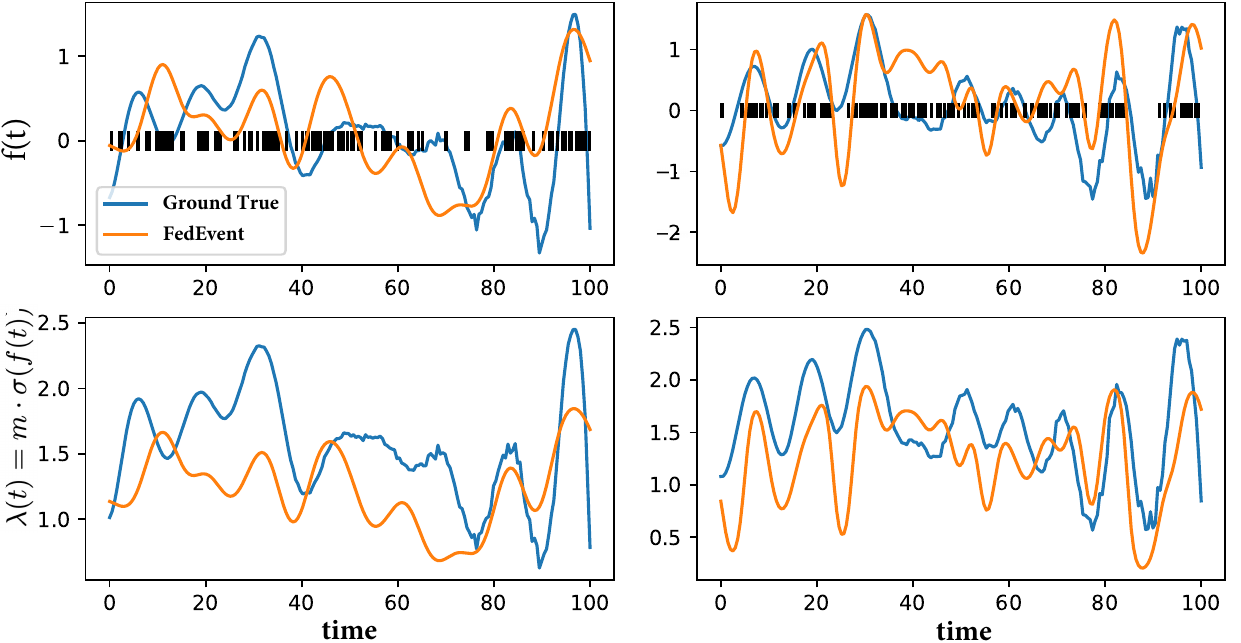}
\caption{Visualisations of models for two clients in the synthetic data  (Client $1$ for the left panel, Client $2$ for the right panel). Top row: Gaussian process random functions $\{f_c(t)\}_{c=1}^2$. Bottom row: intensity functions $\{m_c\cdot\sigma(f_c(t))\}_{c=1}^2$. The short black lines represent the generated events within the time interval. }
\label{fig:simulation-results}
\end{figure}

We use the sparse SGCP to generate synthetic point process data, with the following settings: $m_1=m_c = 50, T = 1$. The RBP kernel hyper-parameters in client 1 and client 2 are: $[1.5, 10], [2,8]$. The number of inducing points is $50$. Fig. \ref{fig:simulation-results} displays the traceplots of fitted random functions against that of groundtruth functions for two clients. 
Overall, our FedPP successfully captures almost all of the complex patterns presented in the groundtruth functions. In the top row, where the random functions $f_c(t)$ are compared, our fitted function closely aligns with the occurrences of events, even outperforming the groundtruth in certain places. For example, within the $[40, 50]$ interval allocated with dense events, the groundtruth function accidentally underestimates these events, whereas our FedPP reflects higher values. In the bottom row, out fitted intensity function is consistently lower than the groundtruth one, particularly during intervals with sparse events.

\subsection{Main Results on Homogeneous Setting}


Apart from the heterogeneous setting, the homogeneous setting where local data across clients are IID is also crucial for testing the effectiveness of federated models. In TPPs tasks, it is natural to construct homogeneous data environments based on different types of events. In this experiment, we evaluate our proposed approach as well as other baselines on each type of events of all datasets. To ensure the validity and fairness of the results, for datasets Taobao, Stack Overflow, and Amazon, we individually select the top 5 event types with the largest sample sizes from each dataset. The numerical results are recorded in the following Table~\ref{tab: ta_ho}-\ref{tab: am_ho}, from which we can see that our model FedPP shows comparable or better results on all datasets. Specifically, FedPP achieves the best results on three event types in the Taobao and two event types in the Stack Overflow and Amazon. For the Conttime dataset, FedPP only perform best in one of the five event types. Moreover, we find that Fed-SAHP, Fed-THP and Fed-Dec-ODE demonstrate strong competitiveness across all datasets.

\begin{table*}[t]
\renewcommand{\arraystretch}{1.3}
\caption{The comparison of test log-likelihood under homogeneous settings on the Taobao dataset.
  }
\label{tab: ta_ho}
\centering
\resizebox{\linewidth}{!}{
\begin{tabular}{cccccc}
\toprule
\multicolumn{6}{c}{Taobao} \\
\midrule 
 Types & Type 1 & Type 2 & Type 3 & Type 4 & Type 5 \\
\midrule
Fed-RMTPPs      & $-2.45\pm0.09$ & $-1.34\pm0.08$ & $-1.22\pm0.07$ & $-2.49\pm0.07$ & $-1.80\pm0.08$   \\ 
Fed-SAHP       & $-0.75\pm0.02$ & $-0.87\pm0.02$  & $-1.24\pm0.11$ & $-1.32\pm0.09$ & $\boldsymbol{-1.68\pm0.03}$  \\
Fed-THP        & $-0.80\pm0.04$ & $\boldsymbol{-0.85\pm0.13}$ & $-1.35\pm0.08$  & $-1.07\pm0.02$ & $-1.72\pm0.06$ \\
Fed-TCVAE  & $\geq -1.23\pm0.07$ & $\geq -1.45\pm0.01$ & $\geq -1.55\pm0.06$ & $\geq -2.03\pm0.12$ & $\geq -2.04\pm0.09$   \\ 
Fed-TCDDM  & $\geq -1.67\pm0.05$ & $\geq -1.48\pm0.03$ & $\geq -1.28\pm0.03$  & $\geq -2.07\pm0.08$ & $\geq -1.83\pm0.04$ \\
Fed-Dec-ODE      & $-0.78\pm0.03$ & $-0.96\pm0.01$  & $-1.30\pm0.06$ & $-0.97\pm0.03$ & $-1.74\pm0.02$ \\
\midrule
FedPP& $\boldsymbol{-0.73\pm0.01}$ & $-0.99\pm0.03$ & $\boldsymbol{-1.19\pm0.01}$ & $\boldsymbol{-0.88\pm0.01}$ & $-1.79\pm0.03$ \\
\bottomrule
\end{tabular}
}
\end{table*} 

\begin{table*}[t]
\renewcommand{\arraystretch}{1.3}
\caption{The comparison of test log-likelihood under homogeneous settings on the Retweet datasets.
  }
\label{tab: re_ho}
\centering
\resizebox{0.7\columnwidth}{!}{
\begin{tabular}{cccc}
\toprule
\multicolumn{4}{c}{Retweet} \\
\midrule 
 Types & Type 1 & Type 2 & Type 3 \\
 \midrule
Fed-RMTPPs      & $-6.34\pm0.07$ & $-6.57\pm0.05$ & $-10.76\pm0.11$    \\ 
Fed-SAHP       & $-4.03\pm0.01$ & $\boldsymbol{-5.23\pm0.02}$  & $-9.27\pm0.04$  \\
Fed-THP        & $-4.12\pm0.06$ & $-5.74\pm0.04$ & $\boldsymbol{-9.03\pm0.09}$  \\
Fed-TCVAE  & $\geq -4.27\pm0.08$ & $\geq -6.74\pm0.12$ & $\geq -10.58\pm0.13$    \\ 
Fed-TCDDM  & $\geq -4.35\pm0.04$ & $\geq -6.55\pm0.07$ & $\geq -10.14\pm0.02$  \\
Fed-Dec-ODE      & $-3.97\pm0.06$ & $-5.88\pm0.04$  & $-9.13\pm0.06$ \\
\midrule
FedPP& $\boldsymbol{-3.27\pm0.03}$ & $-6.35\pm0.02$ & $-10.03\pm0.05$  \\
\bottomrule
\end{tabular}}
\end{table*}

\begin{table*}[t]
\renewcommand{\arraystretch}{1.3}
\caption{The comparison of test log-likelihood under homogeneous settings on the Conttime dataset.
  }
\label{tab: co_ho}
\centering
\resizebox{1\columnwidth}{!}{
\begin{tabular}{cccccc}
\toprule
\multicolumn{6}{c}{Conttime} \\
\midrule 
 Types & Type 1 & Type 2 & Type 3 & Type 4 & Type 5\\
 \midrule
Fed-RMTPPs      & $-2.77\pm0.03$ & $-2.67\pm0.08$ & $-3.01\pm0.08$ & $-3.22\pm0.06$ & $-2.66\pm0.07$   \\ 
Fed-SAHP       & $\boldsymbol{-1.85\pm0.04}$ & $-1.64\pm0.05$  & $\boldsymbol{-2.24\pm0.02}$ & $-2.86\pm0.04$ & $\boldsymbol{-2.45\pm0.03}$  \\
Fed-THP        & $-1.96\pm0.07$ & $-1.77\pm0.03$ & $-2.77\pm0.03$  & $-2.13\pm0.07$ & $-2.73\pm0.09$ \\
Fed-TCVAE  & $\geq -3.02\pm0.01$ & $\geq -2.82\pm0.07$ & $\geq -3.36\pm0.09$ & $\geq -3.77\pm0.15$ & $\geq -3.05\pm0.06$   \\ 
Fed-TCDDM  & $\geq -2.83\pm0.08$ & $\geq -2.93\pm0.06$ & $\geq -2.98\pm0.12$  & $\geq -3.69\pm0.09$ & $\geq -3.72\pm0.04$ \\
Fed-Dec-ODE      & $-1.87\pm0.02$ & $\boldsymbol{-1.58\pm0.03}$  & $-2.68\pm0.05$ & $-2.44\pm0.02$ & $-2.53\pm0.04$ \\
\midrule
FedPP& $-2.17\pm0.08$ & $-2.87\pm0.09$ & $-2.68\pm0.06$ & $\boldsymbol{-2.04\pm0.09}$ & $-2.62\pm0.04$ \\
\bottomrule
\end{tabular}
}
\end{table*}

\begin{table*}[t]
\renewcommand{\arraystretch}{1.3}
\caption{The comparison of test log-likelihood under homogeneous settings on the Stack Overflow dataset.
  }
\label{tab: st_ho}
\centering
\resizebox{1\columnwidth}{!}{
\begin{tabular}{cccccc}
\toprule
\multicolumn{6}{c}{Stack Overflow} \\
\midrule 
 Types & Type 1 & Type 2 & Type 3 & Type 4 & Type 5 \\
\midrule
Fed-RMTPPs      & $-2.88\pm0.07$ & $-3.14\pm0.06$ & $-2.97\pm0.06$ & $-1.99\pm0.08$ & $-2.44\pm0.07$   \\ 
Fed-SAHP       & $-3.01\pm0.02$ & $\boldsymbol{-2.06\pm0.01}$  & $-2.24\pm0.01$ & $-0.45\pm0.02$ & $\boldsymbol{-1.22\pm0.03}$  \\
Fed-THP        & $-2.97\pm0.09$ & $-2.13\pm0.12$ & $-2.37\pm0.08$  & $-0.33\pm0.01$ & $-1.97\pm0.03$ \\
Fed-TCVAE  & $\geq -3.45\pm0.05$ & $\geq -3.55\pm0.15$ & $\geq -3.02\pm0.15$ & $\geq -2.94\pm0.12$ & $\geq -2.53\pm0.17$   \\ 
Fed-TCDDM  & $\geq -3.67\pm0.06$ & $\geq -2.87\pm0.09$ & $\geq -2.96\pm0.11$  & $\geq -1.93\pm0.07$ & $\geq -2.33\pm0.10$ \\
Fed-Dec-ODE      & $-2.28\pm0.03$ & $-2.08\pm0.03$  & $\boldsymbol{-2.17\pm0.03}$ & $-0.42\pm0.05$ & $-1.88\pm0.02$ \\
\midrule
FedPP& $\boldsymbol{-2.17\pm0.09}$ & $-2.09\pm0.01$ & $-2.32\pm0.02$ & $\boldsymbol{-0.08\pm0.01}$ & $-1.73\pm0.02$ \\
\bottomrule
\end{tabular}
}
\end{table*} 

\begin{table*}[t]
\renewcommand{\arraystretch}{1.3}
\caption{The comparison of test log-likelihood under homogeneous settings on the Amazon dataset.
  }
\label{tab: am_ho}
\centering
\resizebox{1\columnwidth}{!}{
\begin{tabular}{cccccc}
\toprule
\multicolumn{6}{c}{Amazon} \\
\midrule 
 Types & Type 1 & Type 2 & Type 3 & Type 4 & Type 5 \\
\midrule
Fed-RMTPPs      & $-1.27\pm0.04$ & $-1.33\pm0.05$ & $-3.16\pm0.18$ & $-1.23\pm0.04$ & $-1.72\pm0.07$   \\ 
Fed-SAHP       & $-0.06\pm0.03$ & $\boldsymbol{0.59\pm0.02}$  & $\boldsymbol{-0.02\pm0.04}$ & $-0.79\pm0.02$ & $-0.62\pm0.04$  \\
Fed-THP        & $0.75\pm0.02$ & $0.47\pm0.08$ & $-2.97\pm0.09$  & $-0.24\pm0.07$ & $-1.12\pm0.13$ \\
Fed-TCVAE  & $\geq -2.13\pm0.14$ & $\geq -0.35\pm0.12$ & $\geq -3.49\pm0.17$ & $\geq -2.34\pm0.13$ & $\geq -1.77\pm0.05$   \\ 
Fed-TCDDM  & $\geq -2.07\pm0.11$ & $\geq -0.87\pm0.13$ & $\geq -3.24\pm0.06$  & $\geq -1.34\pm0.07$ & $\geq -2.34\pm0.14$ \\
Fed-Dec-ODE      & $-0.08\pm0.06$ & $0.33\pm0.04$  & $-015\pm0.07$ & $\boldsymbol{-0.22\pm0.01}$ & $-0.57\pm0.03$ \\
\midrule
FedPP& $\boldsymbol{1.06\pm0.03}$ & $0.57\pm0.16$ & $-0.09\pm0.07$ & $-0.35\pm0.03$ & $\boldsymbol{-0.51\pm0.01}$ \\
\bottomrule
\end{tabular}
}
\end{table*}

\subsection{Main Results on System Heterogeneity}
\begin{figure*}[t]
\centering
\includegraphics[width = 0.32\textwidth]{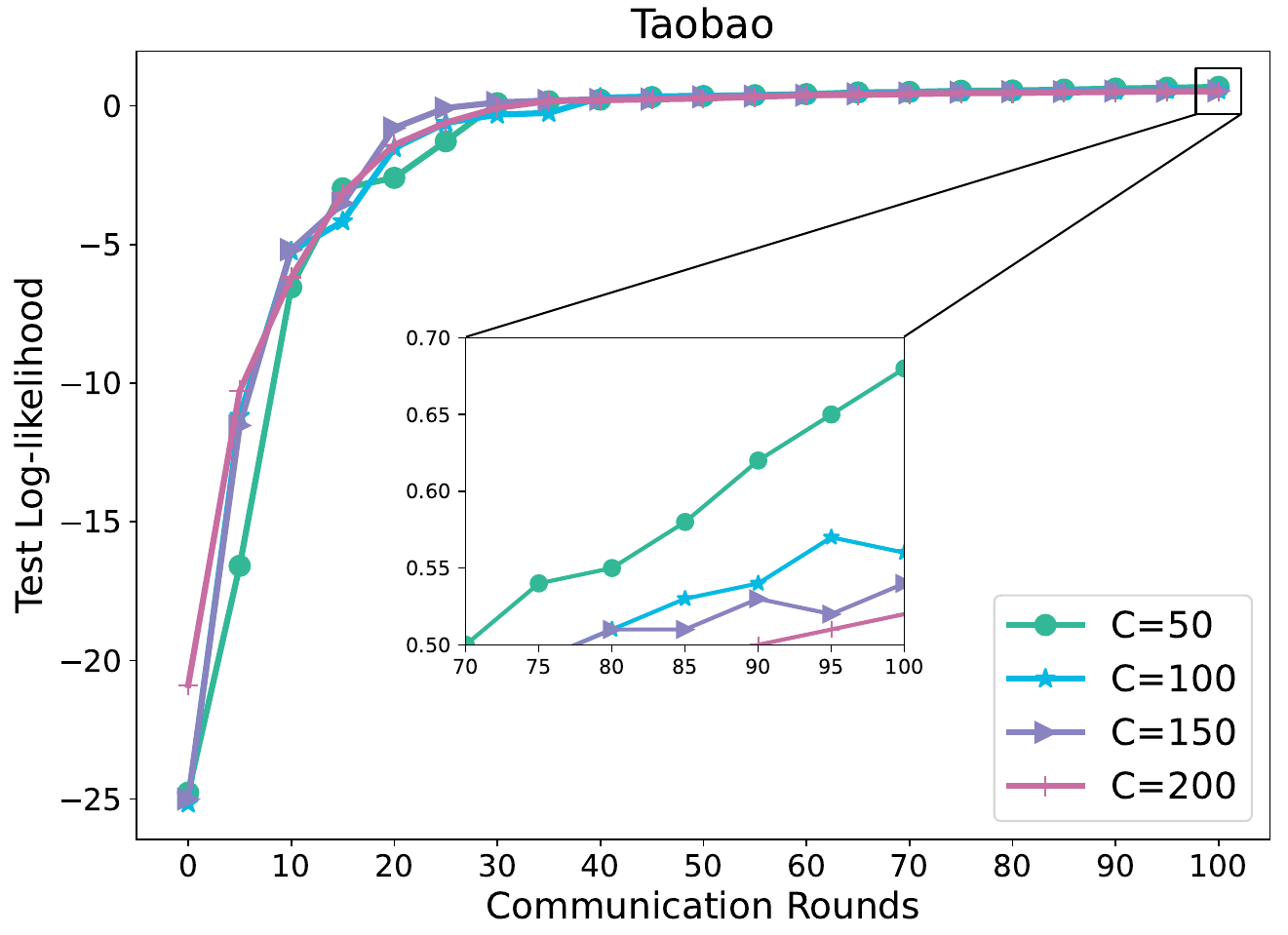}
\includegraphics[width =0.32\textwidth]{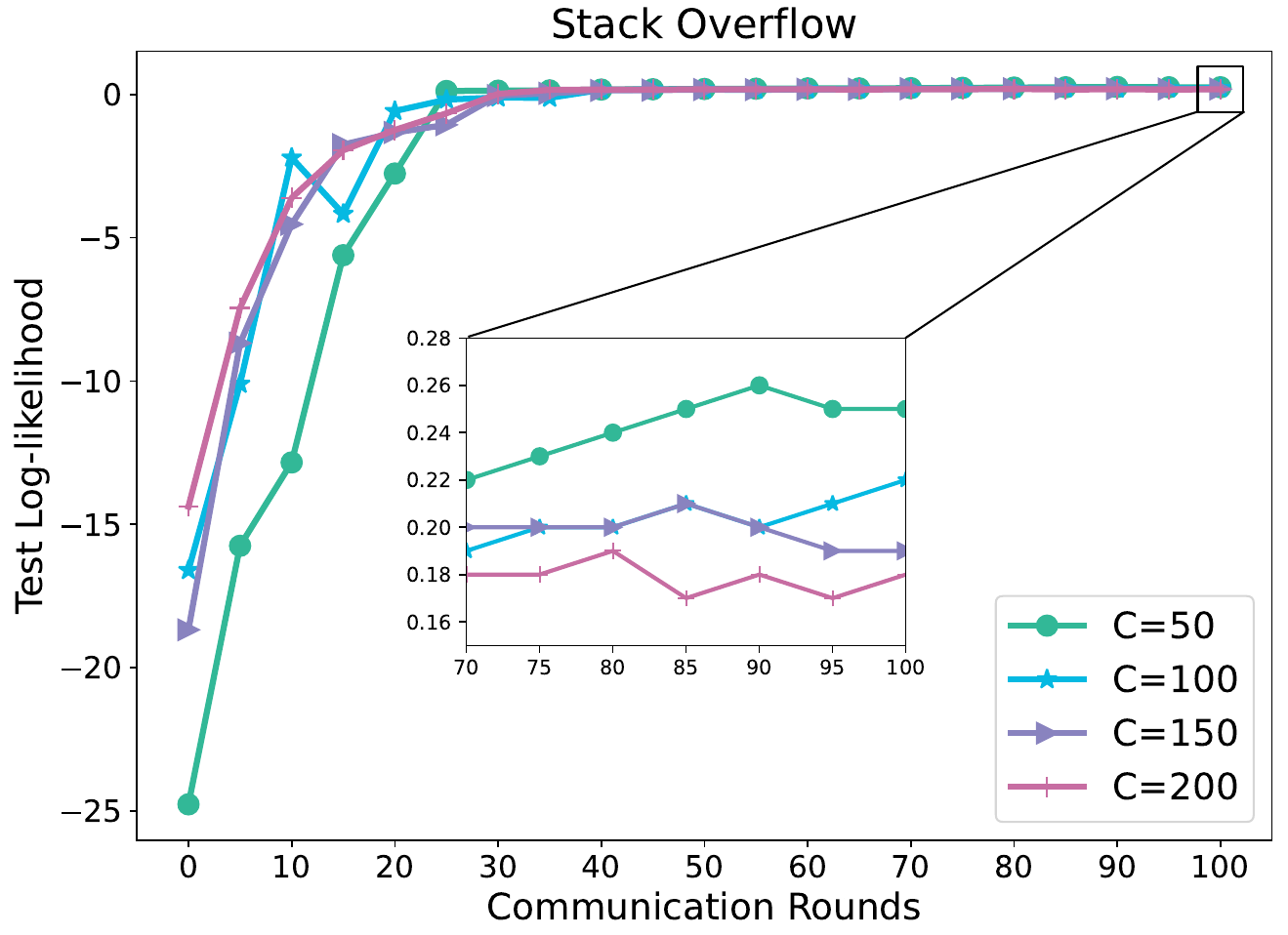}
\includegraphics[width =0.32\textwidth]{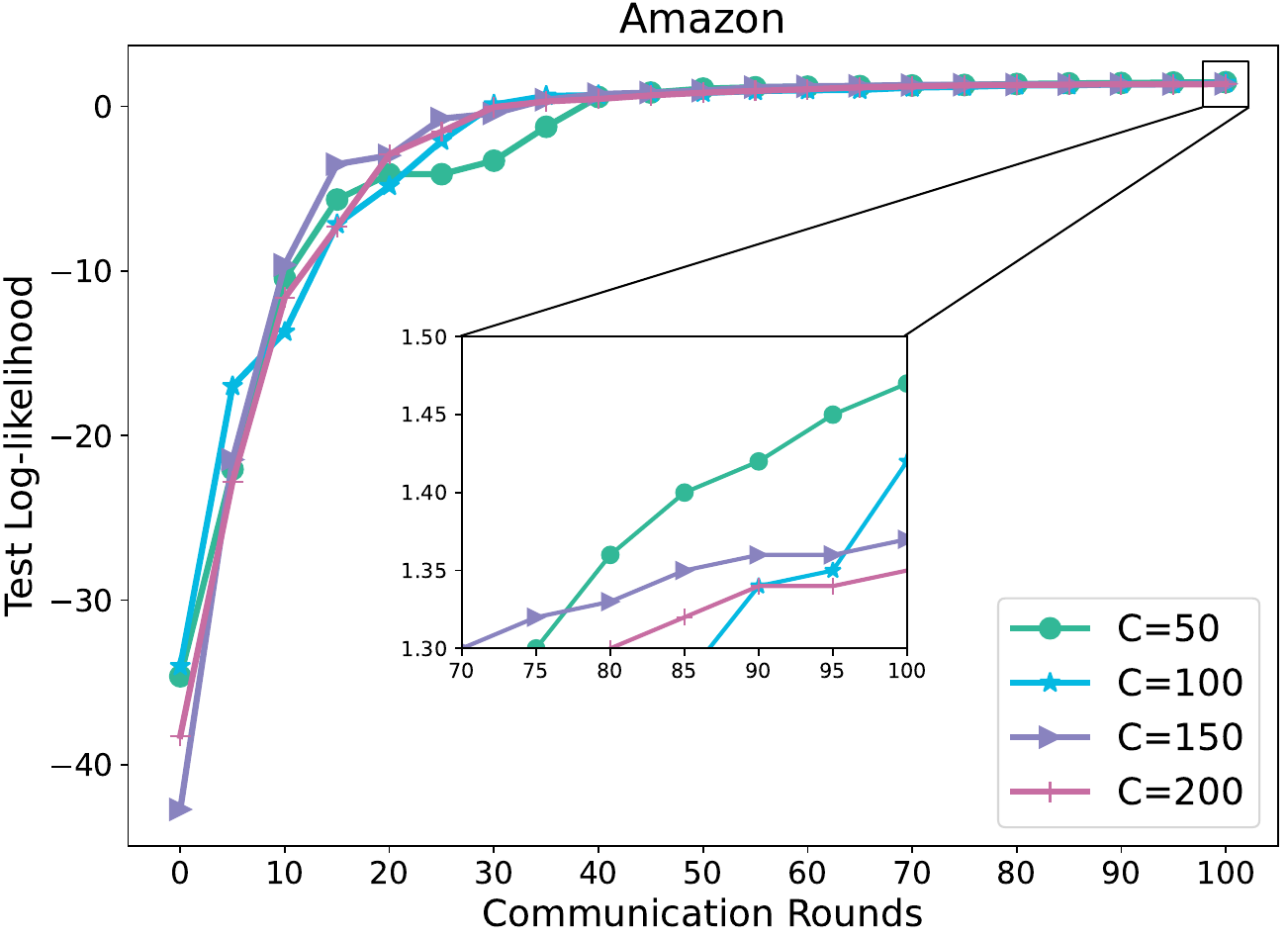}
\caption{Test log-likelihood over communication rounds on Taobao, Stack Overflow and Amazon datasets with varying number of clients.}
\label{fig:client_number}
\end{figure*}

In a practical federated learning setup, where a server collaborates with numerous clients like smartphones, tablets, and laptops. The increase in the number of clients enhances the heterogeneity of the system, resulting in a degradation of model performance. To this end, we evaluate the scalability of our approach on three real-world datasets by adjusting the number of clients $C$ over $\{50, 100, 150, 200\}$. 
To adhere to a cross-device configuration, we set the number of clients participating in each communication round to 10\% of the total number of clients. The convergence results are shown in Fig. \ref{fig:client_number}, from which we can see that as the number of clients increases, the performance of the model across three datasets is deteriorating. This is mainly because, with a constant number of samples, the increase in the number of clients leads to stronger data heterogeneity among different clients, which in turn enlarges the discrepancy between the aggregated global model and the optimal global model. However, it is comforting to note that the degradation in model performance with FedPP is not significant in this scenario, which demonstrates the robustness and practicality of our approach.

\clearpage

\bibliographystyle{elsarticle-num} 
\bibliography{mybibfile}

\begin{thebibliography}{10}
\expandafter\ifx\csname url\endcsname\relax
  \def\url#1{\texttt{#1}}\fi
\expandafter\ifx\csname urlprefix\endcsname\relax\def\urlprefix{URL }\fi
\expandafter\ifx\csname href\endcsname\relax
  \def\href#1#2{#2} \def\path#1{#1}\fi

\bibitem{shchur2021neural}
O.~Shchur, A.~C. T{\"u}rkmen, T.~Januschowski, S.~G{\"u}nnemann, Neural temporal point processes: A review, arXiv preprint arXiv:2104.03528 (2021).

\bibitem{lin2021empirical}
H.~Lin, C.~Tan, L.~Wu, Z.~Gao, S.~Li, et~al., An empirical study: Extensive deep temporal point process, arXiv preprint arXiv:2110.09823 (2021).

\bibitem{wang2024learning}
Z.~Wang, R.~Jiang, H.~Xue, F.~D. Salim, X.~Song, R.~Shibasaki, W.~Hu, S.~Wang, Learning spatio-temporal dynamics on mobility networks for adaptation to open-world events, Artificial Intelligence (2024) 104120.

\bibitem{li2024dual}
X.~Li, Y.~Gong, W.~Liu, Y.~Yin, Y.~Zheng, L.~Nie, Dual-track spatio-temporal learning for urban flow prediction with adaptive normalization, Artificial Intelligence 328 (2024) 104065.

\bibitem{linderman2016bayesian}
S.~W. Linderman, Bayesian methods for discovering structure in neural spike trains, Ph.D. thesis (2016).

\bibitem{apostolopoulou2019mutually}
I.~Apostolopoulou, S.~Linderman, K.~Miller, A.~Dubrawski, Mutually regressive point processes, Advances in Neural Information Processing Systems 32 (2019).

\bibitem{yan2019modeling}
J.~Yan, H.~Xu, L.~Li, Modeling and applications for temporal point processes, in: Proceedings of the 25th ACM SIGKDD International Conference on Knowledge Discovery \& Data Mining, 2019, pp. 3227--3228.

\bibitem{enguehard2020neural}
J.~Enguehard, D.~Busbridge, A.~Bozson, C.~Woodcock, N.~Hammerla, Neural temporal point processes for modelling electronic health records, in: Machine Learning for Health, PMLR, 2020, pp. 85--113.

\bibitem{hawkes2018hawkes}
A.~G. Hawkes, Hawkes processes and their applications to finance: a review, Quantitative Finance 18~(2) (2018) 193--198.

\bibitem{fan2023forecasting}
L.~Fan, H.~Yang, J.~Zhai, X.~Zhang, Forecasting stock volatility during the stock market crash period: The role of {Hawkes} process, Finance Research Letters (2023) 103839.

\bibitem{du2016recurrent}
N.~Du, H.~Dai, R.~Trivedi, U.~Upadhyay, M.~Gomez-Rodriguez, L.~Song, Recurrent marked temporal point processes: Embedding event history to vector, in: Proceedings of the 22nd ACM SIGKDD International Conference on Knowledge Discovery and Data Mining, 2016, pp. 1555--1564.

\bibitem{okawa2019deep}
M.~Okawa, T.~Iwata, T.~Kurashima, Y.~Tanaka, H.~Toda, N.~Ueda, Deep mixture point processes: Spatio-temporal event prediction with rich contextual information, in: Proceedings of the 25th ACM SIGKDD International Conference on Knowledge Discovery \& Data Mining, 2019, pp. 373--383.

\bibitem{mcmahan2017communication}
B.~McMahan, E.~Moore, D.~Ramage, S.~Hampson, B.~A. y~Arcas, Communication-efficient learning of deep networks from decentralized data, in: Artificial Intelligence and Statistics, PMLR, 2017, pp. 1273--1282.

\bibitem{konevcny2016federated}
J.~Kone{\v{c}}n{\`y}, H.~B. McMahan, F.~X. Yu, P.~Richt{\'a}rik, A.~T. Suresh, D.~Bacon, Federated learning: Strategies for improving communication efficiency, arXiv preprint arXiv:1610.05492 (2016).

\bibitem{hu2023privacy}
P.~Hu, Z.~Lin, W.~Pan, Q.~Yang, X.~Peng, Z.~Ming, Privacy-preserving graph convolution network for federated item recommendation, Artificial Intelligence 324 (2023) 103996.

\bibitem{du2024unified}
H.~Du, C.~Cheng, C.~Ni, A unified momentum-based paradigm of decentralized sgd for non-convex models and heterogeneous data, Artificial Intelligence 332 (2024) 104130.

\bibitem{donner2018efficient}
C.~Donner, M.~Opper, Efficient {Bayesian} inference of sigmoidal {Gaussian} {Cox} processes, The Journal of Machine Learning Research 19~(1) (2018) 2710--2743.

\bibitem{zhou2019scalable}
F.~Zhou, Z.~Li, X.~Fan, Y.~Wang, A.~Sowmya, F.~Chen, Efficient inference for nonparametric {Hawkes} processes using auxiliary latent variables, Journal of Machine Learning Research 21~(241) (2020) 1--31.

\bibitem{zhou2020efficient}
F.~Zhou, Y.~Zhang, J.~Zhu, Efficient inference of flexible interaction in spiking-neuron networks, in: International Conference on Learning Representations, 2020.

\bibitem{zuo2020transformer}
S.~Zuo, H.~Jiang, Z.~Li, T.~Zhao, H.~Zha, Transformer {Hawkes} process, in: International Conference on Machine Learning, PMLR, 2020, pp. 11692--11702.

\bibitem{zhang2020self}
Q.~Zhang, A.~Lipani, O.~Kirnap, E.~Yilmaz, Self-attentive {Hawkes} process, in: International Conference on Machine Learning, PMLR, 2020, pp. 11183--11193.

\bibitem{daley2007introduction}
D.~J. Daley, D.~Vere-Jones, An Introduction to the Theory of Point Processes: Volume II: General Theory and Structure, Springer Science \& Business Media, 2007.

\bibitem{xue2023easytpp}
S.~Xue, X.~Shi, Z.~Chu, Y.~Wang, H.~Hao, F.~Zhou, C.~JIANG, C.~Pan, J.~Y. Zhang, Q.~Wen, et~al., Easytpp: Towards open benchmarking temporal point processes, in: The Twelfth International Conference on Learning Representations, 2023.

\bibitem{chen2024fedsi}
H.~Chen, H.~Liu, Z.~Wu, X.~Fan, L.~Cao, Fedsi: Federated subnetwork inference for efficient uncertainty quantification, arXiv preprint arXiv:2404.15657 (2024).

\bibitem{xuhui2021neurips}
X.~Fan, B.~Li, F.~Zhou, S.~Sisson, Continuous-time edge modelling using non-parametric point processes, in: Advances in Neural Information Processing Systems, 2021, pp. 2319--2330.

\bibitem{zhou2023heterogeneous}
F.~Zhou, Q.~Kong, Z.~Deng, F.~He, P.~Cui, J.~Zhu, Heterogeneous multi-task {Gaussian} {Cox} processes, Machine Learning 112~(12) (2023) 5105--5134.

\bibitem{bishop2006pattern}
C.~M. Bishop, Pattern Recognition and Machine Learning (Information Science and Statistics), Springer-Verlag, 2006.

\bibitem{chen2023bayesian}
H.~Chen, H.~Liu, L.~Cao, T.~Zhang, Bayesian personalized federated learning with shared and personalized uncertainty representations, arXiv preprint arXiv:2309.15499 (2023).

\bibitem{DKL}
M.~Patacchiola, J.~Turner, E.~J. Crowley, M.~O'Boyle, A.~J. Storkey, Bayesian meta-learning for the few-shot setting via deep kernels, Advances in Neural Information Processing Systems 33 (2020) 16108--16118.

\bibitem{VB-IB}
M.~K. Titsias, F.~J. Ruiz, S.~Nikoloutsopoulos, A.~Galashov, Information theoretic meta learning with {Gaussian} processes, in: Uncertainty in Artificial Intelligence, PMLR, 2021, pp. 1597--1606.

\bibitem{yu2022federated}
H.~Yu, K.~Guo, M.~Karami, X.~Chen, G.~Zhang, P.~Poupart, Federated {Bayesian} neural regression: A scalable global federated {Gaussian} process, arXiv preprint arXiv:2206.06357 (2022).

\bibitem{achituve2021personalized}
I.~Achituve, A.~Shamsian, A.~Navon, G.~Chechik, E.~Fetaya, Personalized federated learning with {Gaussian} processes, Advances in Neural Information Processing Systems 34 (2021) 8392--8406.

\bibitem{denseGP}
Z.~Wang, Z.~Miao, X.~Zhen, Q.~Qiu, Learning to learn dense {Gaussian} processes for few-shot learning, Advances in Neural Information Processing Systems 34 (2021) 13230--13241.

\bibitem{zhang2022personalized}
X.~Zhang, Y.~Li, W.~Li, K.~Guo, Y.~Shao, Personalized federated learning via variational {Bayesian} inference, in: International Conference on Machine Learning, PMLR, 2022, pp. 26293--26310.

\bibitem{reynaud2013inference}
P.~Reynaud-Bouret, V.~Rivoirard, C.~Tuleau-Malot, Inference of functional connectivity in neurosciences via {Hawkes} processes, in: 2013 IEEE Global Conference on Signal and Information Processing, IEEE, 2013, pp. 317--320.

\bibitem{gerhard2017stability}
F.~Gerhard, M.~Deger, W.~Truccolo, On the stability and dynamics of stochastic spiking neuron models: Nonlinear {Hawkes} process and point process {GLM}s, PLoS Computational Biology 13~(2) (2017) e1005390.

\bibitem{mei2017neural}
H.~Mei, J.~M. Eisner, The neural {Hawkes} process: A neurally self-modulating multivariate point process, Advances in Neural Information Processing Systems 30 (2017).

\bibitem{lin2022exploring}
H.~Lin, L.~Wu, G.~Zhao, L.~Pai, S.~Z. Li, Exploring generative neural temporal point process, Transactions on Machine Learning Research (2022).

\bibitem{song2024decoupled}
Y.~Song, L.~Donghyun, R.~Meng, W.~H. Kim, Decoupled marked temporal point process using neural ordinary differential equations, in: The Twelfth International Conference on Learning Representations, 2024.

\bibitem{kingma2014adam}
D.~P. Kingma, J.~Ba, Adam: A method for stochastic optimization, arXiv preprint arXiv:1412.6980 (2014).

\bibitem{xu2023stabilizing}
J.~Xu, M.~Yang, W.~Ding, S.-L. Huang, Stabilizing and improving federated learning with non-{IID} data and client dropout, arXiv e-prints (2023) arXiv--2303.

\bibitem{jenkins2005survival}
S.~P. Jenkins, Survival analysis, Unpublished manuscript, Institute for Social and Economic Research, University of Essex, Colchester, UK 42 (2005) 54--56.

\bibitem{wang2019machine}
P.~Wang, Y.~Li, C.~K. Reddy, Machine learning for survival analysis: A survey, ACM Computing Surveys (CSUR) 51~(6) (2019) 1--36.

\bibitem{rahimian2022practical}
S.~Rahimian, R.~Kerkouche, I.~Kurth, M.~Fritz, Practical challenges in differentially-private federated survival analysis of medical data, in: Conference on Health, Inference, and Learning, PMLR, 2022, pp. 411--425.

\bibitem{rahman2023fedpseudo}
M.~M. Rahman, S.~Purushotham, Fedpseudo: Privacy-preserving pseudo value-based deep learning models for federated survival analysis, in: Proceedings of the 29th ACM SIGKDD Conference on Knowledge Discovery and Data Mining, 2023, pp. 1999--2009.

\bibitem{rahman2023communication}
M.~M. Rahman, S.~Purushotham, Communication-efficient pseudo value-based random forests for federated survival analysis, in: Proceedings of the AAAI Symposium Series, Vol.~2, 2023, pp. 458--466.

\bibitem{damianou2013deep}
A.~Damianou, N.~D. Lawrence, Deep {Gaussian} processes, in: Artificial intelligence and statistics, PMLR, 2013, pp. 207--215.

\bibitem{wilson2016deep}
A.~G. Wilson, Z.~Hu, R.~Salakhutdinov, E.~P. Xing, Deep kernel learning, in: Artificial Intelligence and Statistics, PMLR, 2016, pp. 370--378.

\bibitem{batz2018approximate}
P.~Batz, A.~Ruttor, M.~Opper, Approximate {Bayes} learning of stochastic differential equations, Physical Review E 98~(2) (2018) 022109.

\bibitem{donner2018efficienta}
C.~Donner, M.~Opper, Efficient {Bayesian} inference for a {Gaussian} process density model, arXiv preprint arXiv:1805.11494 (2018).

\end{thebibliography}

\end{document}